\def\year{2021}\relax
\documentclass[letterpaper]{article} 
\usepackage{aaai21}  
\usepackage{times}  
\usepackage{helvet} 
\usepackage{courier}  
\usepackage[hyphens]{url}  
\usepackage{graphicx} 
\usepackage{amsmath}
\usepackage{amssymb}
\usepackage{multirow}
\usepackage{color}
\usepackage[switch]{lineno}
\usepackage{graphicx}  
\usepackage{natbib}  
\usepackage{caption} 
\title{VAE$^2$: Preventing Posterior Collapse of Variational Video Predictions in the Wild}
\author{
Yizhou Zhou,\textsuperscript{\rm 1}
Chong Luo,\textsuperscript{\rm 2}
Xiaoyan Sun,\textsuperscript{\rm 2}
Zheng-Jun Zha,\textsuperscript{\rm 1}
Wenjun Zeng\textsuperscript{\rm 2} \\ \small
\textsuperscript{\rm 1}University of Science Technology of China ~~~~~~~~~~~~~~~~~~~~~~~~ \textsuperscript{\rm 2}Microsoft Research Asia \\
~~~~~~~~~~~~zyz0205@mail.ustc.edu.cn ~~~ zhazj@ustc.edu.cn ~~~~~~ \{xysun, cluo, wezeng\}@microsoft.com 
}
\affiliations{}

\begin{document}

\maketitle
\begin{abstract}
Predicting future frames of video sequences is challenging due to the complex and stochastic nature of the problem. Video prediction methods based on variational auto-encoders (VAEs) have been a great success, but they require the training data to contain multiple possible futures for an observed video sequence. This is hard to be fulfilled when videos are captured in the wild where any given observation only has a determinate future. As a result, training a vanilla VAE model with these videos inevitably causes posterior collapse. To alleviate this problem, we propose a novel VAE structure, dabbed VAE-in-VAE or VAE$^2$. The key idea is to explicitly introduce stochasticity into the VAE. We treat part of the observed video sequence as a random transition state that bridges its past and future, and maximize the likelihood of a Markov Chain over the video sequence under all possible transition states. A tractable lower bound is proposed for this intractable objective function and an end-to-end optimization algorithm is designed accordingly. VAE$^2$ can mitigate the posterior collapse problem to a large extent, as it breaks the direct dependence between future and observation and does not directly regress the determinate future provided by the training data. We carry out experiments on a large-scale dataset called Cityscapes, which contains videos collected from a number of urban cities. Results show that VAE$^2$ is capable of predicting diverse futures and is more resistant to posterior collapse than the other state-of-the-art VAE-based approaches. We believe that VAE$^2$ is also applicable to other stochastic sequence prediction problems where training data are lack of stochasticity.
\end{abstract}

\section{Introduction}

Video prediction finds many applications in robotics and autonomous driving, such as action recognition\cite{zhou2018mict}, planning\cite{thrun2006stanley}, and object tracking\cite{guo2017learning}. Initially, video prediction was formulated as a reconstruction problem \cite{ranzato2014video} where the trained model regresses a determinate future for any given observation. However, real-world events are full of stochasticity. For example, a person standing may sit down, jump up, or even fall down at the next moment. A deterministic model is not capable of predicting multiple possible futures, but such capability is tremendously desired by intelligent agents as it makes them aware of different possible consequences of their actions in real applications. In order to bring stochasticity into video prediction, methods based on autoregressive models \cite{oord2016pixel}, generative adversarial networks (GANs) \cite{goodfellow2014generative,mirza2014conditional}, and variational auto-encoders (VAEs)\cite{kingma2013auto} have been proposed. Among these, VAE-based methods have received the most attention and they are referred as variational video prediction \cite{babaeizadeh2017stochastic}.

\begin{figure*}
    \centering
    \includegraphics[width=0.8\textwidth]{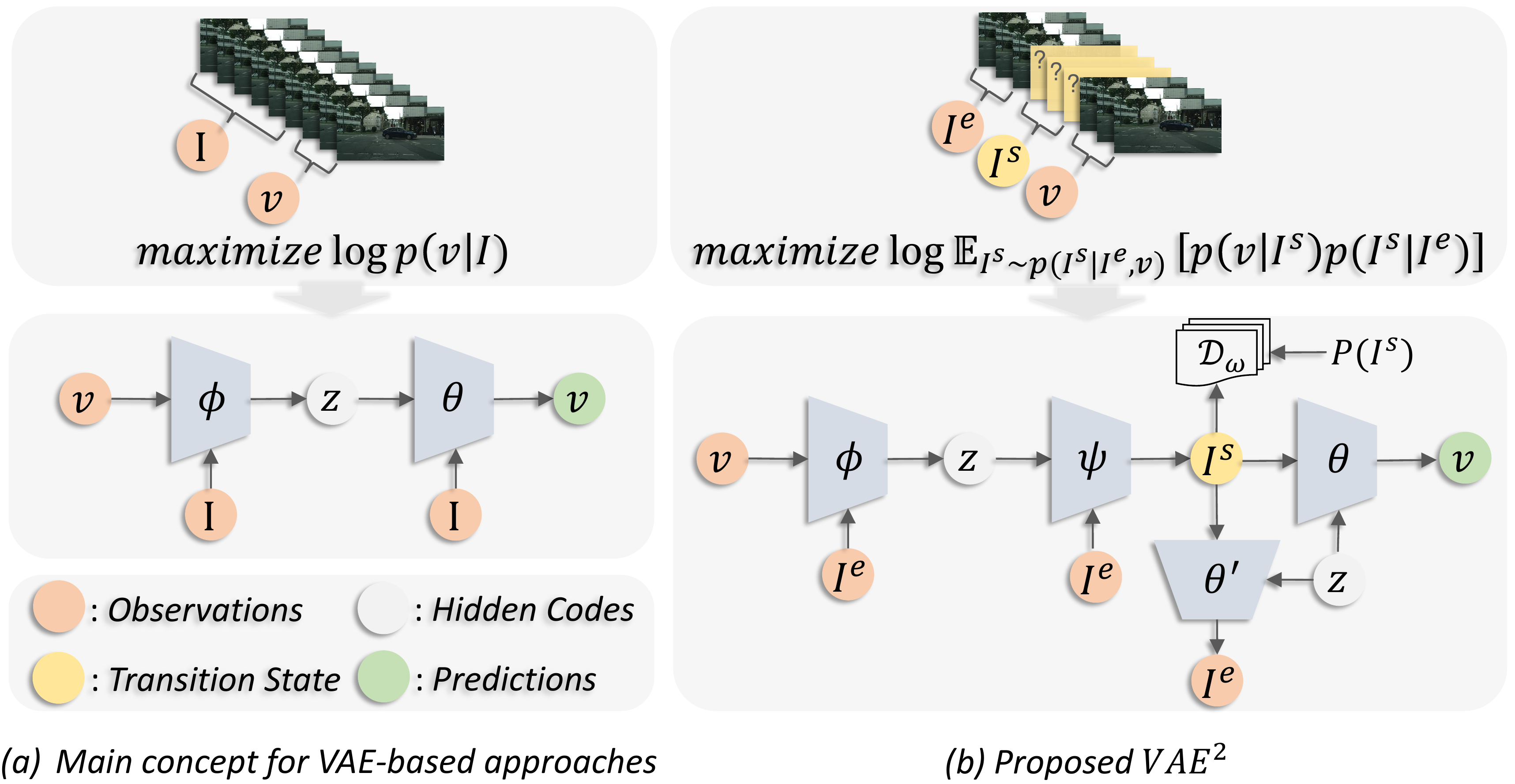}
    \caption{Schematic diagrams of vanilla VAE-based video prediction approach and our VAE$^2$ approach. $\phi$ is the encoder that maps the observation $I$ and future $v$ to a hidden random variable $z$. $\theta$ is the decoder which predicts stochastic future based on the observation and random signal sampled from $z$. When the future is determinate, the decoder $\theta$ can well reconstruct the future without accessing the random hidden code $z$. As a consequence, it leads VAE to a deterministic model. Unlike vanilla VAE that maximizes the log likelihood of determinate pair ($v$, $I$), we propose to treat part of the observation as an unknown transition state (frames $I^s$), and maximize the likelihood of the Markov Chain under all possible transition state. Such formulation breaks the direct dependence between future and observation, so that our method models a stochastic process. A nested VAE structure is then derived for end-to-end optimization, where $\psi$ and $\theta^{'}$ are two additional decoder to produce the transition frames and reconstruct the observed frames, respectively. $D_\omega$ is a discriminator to help generate more realistic transitiona state frames.}
    \label{mainConcept}
\end{figure*}

Variational video prediction learns a latent variable model that maximizes the likelihood of the data. A key to the success of variational video prediction is that the training dataset should provide multiple futures for observed frames. Not surprisingly, VAE-based video prediction methods have been using synthetic videos or scripted videos\footnote{A human actor or robot conducts predefined activities in well-controlled environment.} for training. These videos can provide multiple futures as desired, but they only cover a small subset of real-world scenarios. In order to build a practically applicable video prediction model, it is necessary to train it with non-scripted real-world videos such as videos captured in the wild. However, such kind of videos are usually determinate, which means only one of many possible futures is available. This situation will easily collapse a VAE model. As Fig. \ref{mainConcept}(a) shows, if there is always a unique future $v$ corresponding to each observation $I$, the hidden code $z$ becomes trivial due to the inference preference property\cite{chen2016variational}. In such a case, VAE loses the capability to predict diverse futures.

VAEs are known to suffer from posterior collapse due to various reasons such as the `optimization challenges'\cite{bowman2015generating,razavi2019preventing}. Although many approaches \cite{higgins2017beta,alemi2017fixing,goyal2017z,razavi2019preventing,bowman2015generating} have been proposed to mitigate the problem, they hold the basic assumption that the training data have a stochastic nature. To our best knowledge, none of the previous work has looked into the model collapse problem caused by the determinate training data in video prediction. The intuition behind our solution is to explicitly inject stochasticity into the VAEs. As illustrated in Fig. \ref{mainConcept}(b), we intentionally set aside a part of the observed video sequence $I^s$ and treat it as a random transition state that bridges its past and future. By doing so, we can maximize the likelihood of a Markov Chain over the sequence under all possible transition states. Such formulation converts the likelihood that only relies on the observed data pairs ($I, v$) into an expectation which contains extra dependence on the distribution of the entire dataset. 

However, this new formulation contains an expectation term and a likelihood term which are both intractable. To tackle it, we first derive a practical lower bound which can be optimized with a vanilla VAE structure, and then use another VAE to approximate the remaining intractable part in this lower bound. In addition, we find that the two objective functions for the two VAEs can be merged into a single expression in practice. This greatly saves the efforts to do iterative optimization. As a result, we can innovatively derive a nested VAE structure. We name this structure VAE-in-VAE, or VAE$^2$ in short. VAE$^2$ can be optimized in an end-to-end manner. 

In a nutshell, the contributions of this work are: 1) We propose a novel VAE$^2$ framework to explicitly inject stochasticity into the VAE to mitigate the posterior collapse problem caused by determinate training data. 2) We turn the objective function tractable and develop an efficient end-to-end optimization algorithm. 3) We make evaluations on non-scripted real-world video prediction as well as a simple number sequence prediction task. Both quantitative and qualitative results demonstrate the efficacy of our method.

\section{Related Work}

The video prediction problem can be addressed by either deterministic or non-deterministic models.
Deterministic models directly reconstruct future frames with recurrent neural networks\cite{ranzato2014video,oh2015action,srivastava2015unsupervised, villegas2017decomposing,finn2016unsupervised,lu2017flexible} or feed-forward networks\cite{jia2016dynamic,vondrick2017generating,liu2017video,walker2015dense}. The reconstruction loss assumes a deterministic environment. Such models cannot capture the stochastic nature in real world videos and 
usually result in an averaged prediction over all possibilities.

Non-deterministic models can be further classified into autoregressive models, GANs, and VAEs. In pixel-level autoregressive models\cite{oord2016pixel}, spatiotemporal dependencies are jointly modeled, where each pixel in the predicted video frames fully depends on the previously predicted pixel via chain rule \cite{kalchbrenner2017video}. Although autoregressive model can directly construct video likelihood through full factorization over each pixel, it is unpractical due to its high inference complexity. Besides, it has been observed to fail on globally coherent structures\cite{razavi2019preventing} and generate very noisy predictions \cite{babaeizadeh2017stochastic}. GANs\cite{goodfellow2014generative} and conditional GANs (c-GANs)\cite{mirza2014conditional} are also employed for stochastic video prediction, for their capability to generate data close to the target distribution. GAN-based approaches can predict sharp and realistic video sequences\cite{vondrick2016generating}, but is prone to model collapse and often fails to produce diverse futures\cite{lee2018stochastic}.

VAE-based models have received the most attention among the non-deterministic models. In particular, conditional VAEs (c-VAEs) have been shown to be able to forecast diverse future actions a single static image \cite{walker2016uncertain,xue2016visual}. c-VAEs have also been used to predict diverse future sequences from an observed video sequence \cite{babaeizadeh2017stochastic,lee2018stochastic,denton2018stochastic,minderer2019unsupervised}. In some of these approaches, human body skeleton\cite{minderer2019unsupervised} and dense trajectory\cite{walker2016uncertain} are incorporated in addition to the RGB frames to enhance the prediction quality. Besides, RNN-based encoder-decoder structures such as LSTM\cite{hochreiter1997long} has also been employed for long-term predictions\cite{wichers2018hierarchical}. Some other works leverage GANs to further boost the visual quality\cite{lee2018stochastic}.

So far, the success of VAE-based video prediction methods has been limited to synthetic videos or scripted videos, such as video games or synthetic shapes rendered with multiple futures\cite{xue2016visual,babaeizadeh2017stochastic}, BAIR robotic pushing dataset\cite{ebert2017self} with randomly moving robotic arms that conduct multiple possible movements\cite{babaeizadeh2017stochastic,lee2018stochastic}, or KTH\cite{schuldt2004recognizing} and Human3.6M\cite{ionescu2013human3} where one volunteer repeatedly conducts predefined activities (such as hand clapping and walking) in well controlled environments\cite{lee2018stochastic,denton2018stochastic,babaeizadeh2017stochastic,minderer2019unsupervised}. 
In these datasets, multiple futures are created for a given observation, so that the VAE model can be trained as desired. However, videos captured in the wild are usually determinate. There is always a unique future for a given observation. Such a data distribution can easily result in posterior collapse and degenerate VAE to a deterministic model. A possible fix is to treat future frames at multiple time steps as multiple futures at a single time step\cite{xue2016visual,walker2016uncertain}, but such manipulation creates a non-negligible gap between the distributions of the training data and the real-world data. 
In this paper, we propose VAE$^2$ to alleviate the posterior collapse problem in variational video prediction. Different from previous attempts that address model collapse in VAEs such as employing weak decoder\cite{bowman2015generating,gulrajani2016pixelvae}, involving stronger constraints\cite{higgins2017beta,goyal2017z} or annealing strategy\cite{kim2018semi,gulrajani2016pixelvae,bowman2015generating}, VAE$^2$ is specially designed to handle the collapse problem caused by videos lacking of stochasticity.

\section{VAE-in-VAE}
In this section, we elaborate the proposed VAE$^2$ step by step. We start with introducing the VAE's posterior collapse in video prediction caused by the determinate data pair. Next, we describe how to overcome this problem by injecting stochasticity into vanilla VAE's objective function. Then, we propose a tractable lower bound to facilitate gradient-based solution and finally derive the VAE-in-VAE structure for end-to-end optimization.

\subsection{Posterior Collapse of Vanilla VAE with Determinate Data Pair}
We first briefly introduce the formulation of traditional VAE-based solutions for video predictions.
Let $\mathcal{D} = \{(I_k, v_k)\}$ denote some i.i.d. data pairs consisting of $K$ samples. Assuming $v$ is conditioned on $I$ with some random processes involving an unobserved hidden variable $z$, one can maximize the conditional data likelihood $\sum_{k=1}^{K}\log p(v_k \mid I_k)$ with conditioned VAEs (c-VAEs) through maximizing the variational lower bound
\begin{equation}
    \mathbb{E}_{q_\phi(z_k \mid I_k, v_k)}[p_\theta(v_k \mid I_k, z_k)] - KL(q_\phi(z_k \mid I_k, v_k) \mid\mid p(z)),
\label{vae}
\end{equation}
where $q_\phi$ is a parametric model that approximates the true posterior $p(z_k \mid I_k, v_k)$, $p_\theta$ is a generative model w.r.t. the hidden code $z_k$ and the data $I_k$. $KL$ denotes the Kullback-Leibler(KL) Divergence. 

When VAE is used for stochastic video prediction, each $I_k \in \mathbb{R}^{T_I \times H \times W}$ represents an observed video sequence consisting of $T_I$ consecutive frames with $H \times W$ spatial size, $v_k \in \mathbb{R}^{T_v \times H \times W}$ is the future sequence of the observation consisting of $T_v$ consecutive frames. A general framework for variational video prediction is illustrated in Fig. \ref{mainConcept} (a), where an encoder $\phi$ and a decoder $\theta$ are employed to instantiate $q_\phi$ and $p_\theta$, respectively.

In Eq. \ref{vae}, there is a regression term $p_\theta(v_k \mid I_k, z_k)$ which is usually modeled by a deep decoder that takes as inputs the observation $I_k$ and hidden code $z_k$ and directly regresses the future $v_k$.
Since the videos in the wild are determinately captured, each $I_k$ is only associated with a unique $v_k$ without stochasticity. In principle, such determinate data pair $(I_k, v_k)$ can be easily fit by the decoder $\theta$ since networks with sufficient capacity are capable of representing arbitrarily complex functions\cite{hornik1989multilayer} or even memorize samples\cite{arpit2017closer}. Therefore, the hidden code $z$ can be entirely ignored by the decoder to fulfill the KL divergence term based on the information preference property\cite{chen2016variational}. As a result, VAE is modeling a deterministic process under this scenario.

\subsection{From VAE to VAE$^2$ by Introducing Stochasticity}
The reason for the aforementioned collapse issue is that traditional c-VAEs maximize the likelihood $p(v_k \mid I_k)$ over data pair $(v_k, i_k)$ which has no stochastic behavior. The key to avoid such collapse is to ensure that the VAE is modeling a stochastic process instead of a deterministic one. To achieve this, we split each observation $I_k$ into two parts: determinate observation $I^e_k \in \mathbb{R}^{\frac{T_I}{2} \times H \times W}$ and random transition state $I^s_k \in \mathbb{R}^{\frac{T_I}{2} \times H \times W}$. Different from the determinate $I^e_k$, $I^s_k$ is treated as a random event that bridges the determinate observation and the unknown future. Assuming that evolution of a video sequence subjects to a Markov process, where the generation process of a sequence is only conditioned on its previous sequence, we propose to maximize the likelihood of the Markov Chain over observations and futures under all possible transition states. The optimization problem can be expressed as
\begin{equation}
    maximize~ \log \mathbb{E}_{p(I^s_k \mid v_k, I^e_k)} [p(v_k \mid I^s_k) p(I^s_k \mid I^e_k)].
\label{OriginalobjectiveFunction}    
\end{equation}
Intuitively, the proposed objective function involves stochastic information by explicitly relaxing a part of determinate observations to random transition state. However, such formulation is intractable in terms of both the likelihood of the Markov Chain and the expectation term.

\subsection{A Tractable Objective Function for VAE$^2$}
In this section, we demonstrate that a tractable lower bound for Eq. \ref{OriginalobjectiveFunction} can be derived by applying Cauchy-Schwarz inequality. Specifically, we have
\begin{equation}
\begin{split}
    &\log \mathbb{E}_{p(I^s \mid v, I^e)} [p(v \mid I^s) p(I^s \mid I^e)] \\
    &\geq  \mathbb{E}_{q_\phi(z \mid I^e, v)}[\log \mathbb{E}_{p(I^s \mid I^e, z)}p_\theta(v \mid I^s, z)] \\
    &~~~~~~~~- KL(q_\phi(z \mid I^e, v) \mid\mid p(z)). \\
\end{split}
\label{lowerBound}
\end{equation}
Here we omit index $k$ for simplicity. The above lower bound serves as our first-level objective function. It has a similar form as Eq. \ref{vae} except that we have one additional transition variable $I^s$ in the decoder model $p_\theta$. To maximize this lower bound, we need to calculate the expectation term $\mathbb{E}_{p(I^s \mid I^e, z)}p_\theta(v \mid I^s, z)$ in Eq. \ref{lowerBound}, which requires an approximation towards $p(I^s \mid I^e, z)$. Here, we employ a generation model $q_\psi(I^s \mid I^e, z)$ parameterized with $\psi$ for the approximation by minimizing $KL(q_\psi(I^s \mid I^e, z) \mid\mid p(I^s \mid I^e, z))$, which induces our second-level objective function
\begin{equation}
    \mathbb{E}_{q_\psi(I^s \mid I^e, z)}\log p_{\theta^{'}}(I^e \mid I^s, z) - KL(q_\psi(I^s \mid I^e, z) \mid\mid p(I^s)).
\label{secondLowerBound}
\end{equation}
For the full derivation of Eq. \ref{lowerBound} and Eq. \ref{secondLowerBound}, \textbf{please refer to our appendix}. Since transition state $I^s$ is high-dimensional signals (video sequences), the exact functional form of its prior distribution is not accessible. We follow adversarial autoencoders\cite{makhzani2015adversarial} to leverage adversarial training to minimize the distance between the $q_\psi(I^s \mid I^e, z)$ and the prior $p(I^s)$. As such, $-KL(q_\psi(I^s \mid I^e, z) \mid\mid p(I^s))$ in Eq. \ref{secondLowerBound} is replaced by
\begin{equation}
    D_\omega(I^s, I_{prior}), ~where~ I_{prior} \sim p(I^s), I^s \sim q_\psi(I^s \mid I^e, z).
\end{equation}
Here, $D_\omega$ is a discriminator network parameterized with $\omega$ and $I_{prior}$
is randomly sampled from the video dataset. By doing so, the second-level objective function Eq. \ref{secondLowerBound} is now tractable.
Ideally, we shall optimize the first-level and second-level objective function iteratively, which requires unaffordable training time for convergence. In practice, we simplify such iterative learning process by merging the two objective functions into a single one as
\begin{equation}
\begin{split}
    &\mathcal{L}(I^e, I^s, v; \theta, \phi) + \\
    & \lambda [\mathbb{E}_{q_\psi(I^s \mid I^e, z)}\log p_{\theta^{'}}(I^e \mid I^s, z) + {D}_\omega(I^s, I_{prior})],
\end{split}
\label{objectiveFunction}
\end{equation}
where $\lambda$ is a loss weight applied on Eq. \ref{secondLowerBound}. This simplification enables us to simultaneously optimize the two objective functions and we use a weight $\lambda$ to adjust the optimization speed of our second-level objective function to mimic the original iterative process. 

\subsection{End-to-end Optimization}
We employ two VAE structures to maximize our final objective function in Eq. \ref{objectiveFunction}. As illustrated in Fig. \ref{mainConcept}(b), the first VAE consisting of an encoder $\phi$ and a decoder $\theta$ is incorporated to maximize the first-level objective function $\mathcal{L}(I^e, I^s, v; \theta, \phi)$. The second VAE with an encoder $\psi$, a decoder ${\theta^{'}}$ and a discriminator $D_\omega$ is used for maximizing the second-level objective function. 
We assume $p_\theta$ and $p_{\theta^{'}}$ to be Laplace distribution (which leads to L1 regression loss) and assume $q_\phi$ to be Gaussian. The training procedure can be summarized as follows:
\begin{itemize}
    \item Encoder $\phi$ takes $I^e$ and $v$ as inputs and produces hidden codes $z$. $z$ is fed into $\psi$ to generate $L$ different $I^s$. For each $I^s$, decoder $\theta$ and $\theta^{'}$ reconstruct $v$ and $I^e$, respectively.
    \item Estimate the $\log\mathbf{E}_{p(I^s \mid I^e, z)}p_\theta(v \mid I^s, z)$ in Eq. \ref{lowerBound} with $\log \sum_{i=1}^{L} p_\theta(v \mid I^s_i, z)$. Estimate $\mathbf{E}_{q_\psi(I^s \mid I^e, z)}\log p_{\theta^{'}}(I^e \mid I^s, z)$ in Eq. \ref{objectiveFunction} with $\sum_{i=1}^{L} \log p_{\theta^{'}}(I^e \mid I^s_i, z)$.
    \item Compute gradients w.r.t. Eq. \ref{objectiveFunction} and update $\phi$, $\psi$, $\theta$, $\theta^{'}$. Update $D_\omega$ with adversarial learning.
\end{itemize}
In practice, we observe that this simplified process works well even with $L=1$. This further reduces the algorithm complexity.

\section{Number Sequence Prediction}
\begin{figure*}
    \centering
    \includegraphics[width=0.95\textwidth]{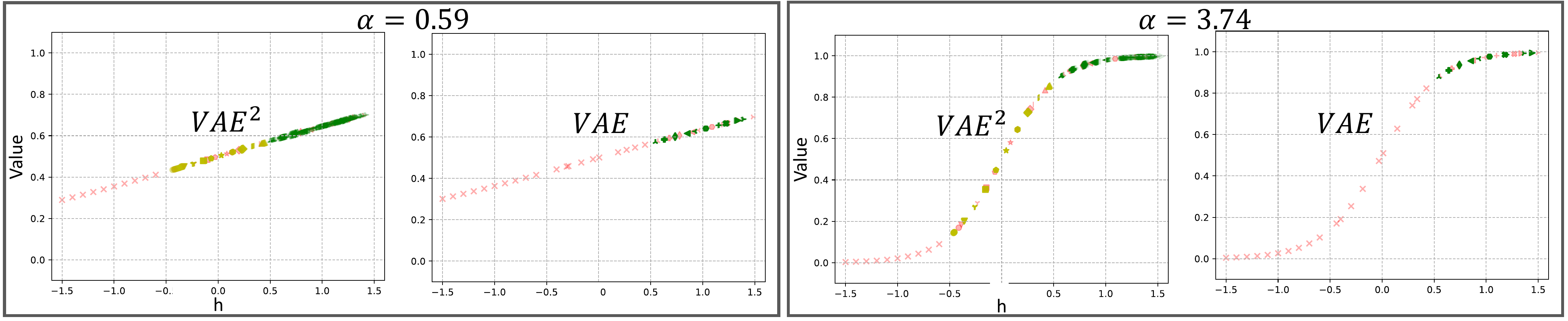}
    \caption{Visualization of the number sequence prediction at two model parameters. The 100 predictions made by VAE baseline are almost identical while those made by the proposed VAE$^2$ are more scattered around the groundtruth. This shows that VAE$^2$ captures the inherent stochastic law of the number sequences model.}
    \label{toyEcxample}
\end{figure*}
We first use a simple number sequence prediction task to demonstrate how VAE$^2$ mitigates the collapse problem when only determinate data are available. More specifically, we explicitly design a world model which can stochastically produce number sequences w.r.t. the given model parameter. 

\textbf{Sequence Model Design}.
The stochastic model that generates number sequence is defined by $G(\alpha,H(\boldsymbol{\epsilon})) = \frac{1}{1+e^{-\alpha H(\boldsymbol{\epsilon})}}$. Here, $H(\boldsymbol{\epsilon}) = [h_0(\epsilon_0), h_1(\epsilon_1), ..., h_{N-1}(\epsilon_{N-1})] \in \mathbb{R}^N$ and $h_i(\epsilon_i) = C+i*S+\epsilon_i$, where $\epsilon_i \sim Uniform(0, S)$.
This world model is parameterized by $\alpha$ and its stochastic nature is characterized by $H(\boldsymbol{\epsilon})$. 

\textbf{Constructing the Dataset}.
In order to mimic the determinate video sequences captured in the wild, we use $G(\alpha,H(\epsilon))$ to generate only one number sequence for each world parameter $\alpha$. We then split each number sequence into three even parts. For VAE$^2$, the three parts correspond to $I^e$, $I^s$ and $v$, respectively. For the baseline VAE, the first and the second parts together correspond to $I$ and the third part corresponds to $v$. The dataset $D = \{G(\alpha_k, H(\boldsymbol{\hat\epsilon})), k\in[1,K]\}$ contains $K$ number sequences generated from $K$ different world model parameters, where $H(\boldsymbol{\hat\epsilon})$ denotes a single sampling result of $H(\boldsymbol{\epsilon)}$. 
We set $C$, $S$, $N$, and $K$ to -1.5, 0.1, 30 and 10,000, respectively. The constructed dataset contains 10,000 number sequences, each of which has 30 data points.

\textbf{Evaluation and Visualization}.
We train our VAE$^2$, VAE\cite{babaeizadeh2017stochastic} and VAE-GAN\cite{lee2018stochastic} models on this number sequence dataset, respectively. The training and architecture details can be found in the appendix. After training, we make 100 predictions on $v$ using each method. In Fig. \ref{toyEcxample}, we plot original data points in red and predicted ones in green. The different shapes correspond to different hidden variables $h_i(\epsilon_i)$. 

We observe from the figure that the predicted $v$ from the baseline VAE model are almost identical among the 100 predictions (samples of the same shape are grouped together), showing that the baseline VAE degrades to a deterministic model. In contrast, the proposed VAE$^2$ provides much more diverse predictions (samples of each shape scatter around the ground-truth). Although we do not provide multiple futures in the training dataset, VAE$^2$ is able to explore the underlying stochastic information through our innovative design. More visualizations can be found in the appendix.

In addition to visualizing the predicted numbers, we can use the standard deviation of the L1 loss of different samples to measure the diversity of predictions. Fig. \ref{cityscapes-variance}(a) plots the mean and the standard deviation of the predictions from VAE$^2$ and its counterparts on the entire dataset. It is clear that the number sequences predicted by the proposed VAE$^2$ are more diverse than other methods. 

\section{Experiments on Videos Captured in the Wild}
\begin{figure*}
    \centering
    \includegraphics[width=0.95\textwidth]{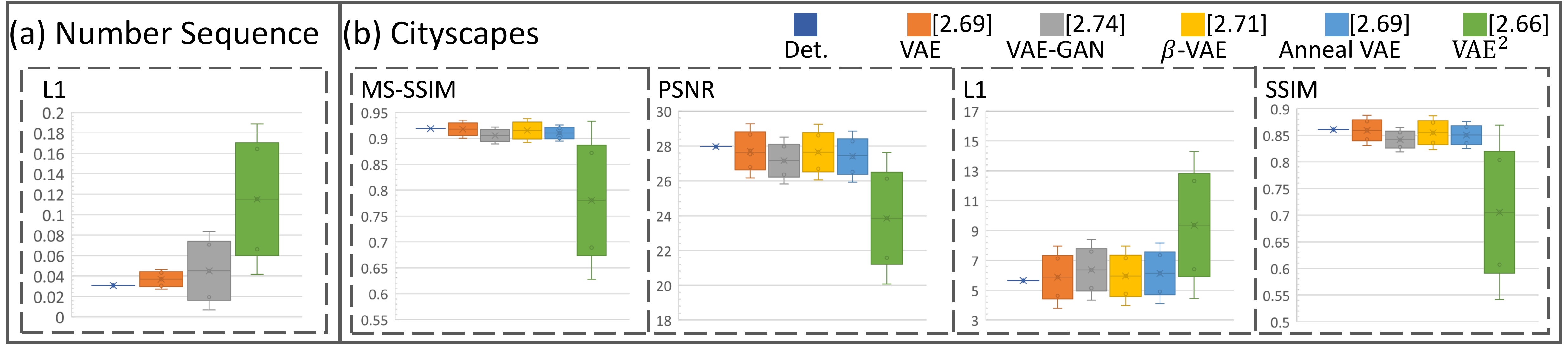}
    \caption{Standard deviation of 100 prediction samples from different methods under four criteria. Each box and whisker indicates the mean $\pm$ three times and five times standard deviation, respectively. Larger deviation range of VAE$^2$ indicates it can significantly improve the diversity of the predictions. The number beside each legend denotes the averaged Inception Score of the 100 samples from corresponding method over the entire Cityscapes dataset.}
    \label{cityscapes-variance}
\end{figure*}
\textbf{Dataset and Evaluation Metrics}.
We evaluate the proposed VAE$^2$ with the Cityscapes dataset\cite{cordts2016cityscapes} which contains urban street scenes from 50 different cities. The videos are captured at 17 fps with cameras mounted on moving cars. Since a car and its mounted camera pass each street only once, every video sequence is determinate and there are no multiple futures available. The average number of humans and vehicles per frame is 7.0 and 11.8, respectively, providing a fairly complex instance pattern. 

There is no consensus yet on how to evaluate a video prediction scheme. Previous work has tried to evaluate the perceived visual quality of the predicted frames. However, this work is focused on mitigating the model collapse problem caused by determinate training data. In addition to the perceived visual quality, we will mainly evaluate how seriously a prediction model suffers from the model collapse problem. In general, the more diverse the predicted frames are, the better the model handles the model collapse problem. The diversity of predicted frames can be evaluated both quantitatively and qualitatively. In particular, we use the standard deviation of some conventional image quality evaluation metrics, such as SSIM\cite{wang2004image}, PSNR\cite{huynh2008scope}, and L1 loss, for quantitative evaluation. We also compute the optical flows between the ground-truth future frame and the predicted future frames by various prediction models. Since optical flow reflects per-pixel displacement, it can be a very intuitive way to show the pixel-level difference between the two frames.

\textbf{Reference Schemes and Implementation Details}.
Existing VAE-based video prediction approaches are designed with different backbone networks and data structures, including RGB frames, skeleton, or dense trajectory. In this work, we only consider methods that directly operate on raw RGB videos to demonstrate the efficacy of the proposed VAE$^2$. These prediction models can be categorized into two VAE variants, namely vanilla VAEs \cite{babaeizadeh2017stochastic,xue2016visual,denton2018stochastic} and VAE-GAN \cite{lee2018stochastic,larsen2015autoencoding}. In addition, we evaluate $\beta$-VAE\cite{higgins2017beta} and VAE with annealing\cite{bowman2015generating} which are designed for addressing the model collapse problem without considering the situation where training data lack of stochasticity. We also construct a deterministic model as a baseline.

We employ 18-layer HRNet (wd-18-small)\cite{sun2019deep} to instantiate all the encoders, decoders, and GANs in VAE$^2$. For fair comparison, we replace the original backbone network in the reference schemes \cite{babaeizadeh2017stochastic,lee2018stochastic,higgins2017beta,bowman2015generating} with the same HRNet. The deterministic baseline is also a 18-layer HRNet which directly regresses the future frames. 
We set $\lambda$ to 0.1 and we use Adam optimizer to train all methods for 1,000 epochs. More training details can be found in the appendix.

\begin{figure}
    \centering
    \includegraphics[width=0.9\linewidth]{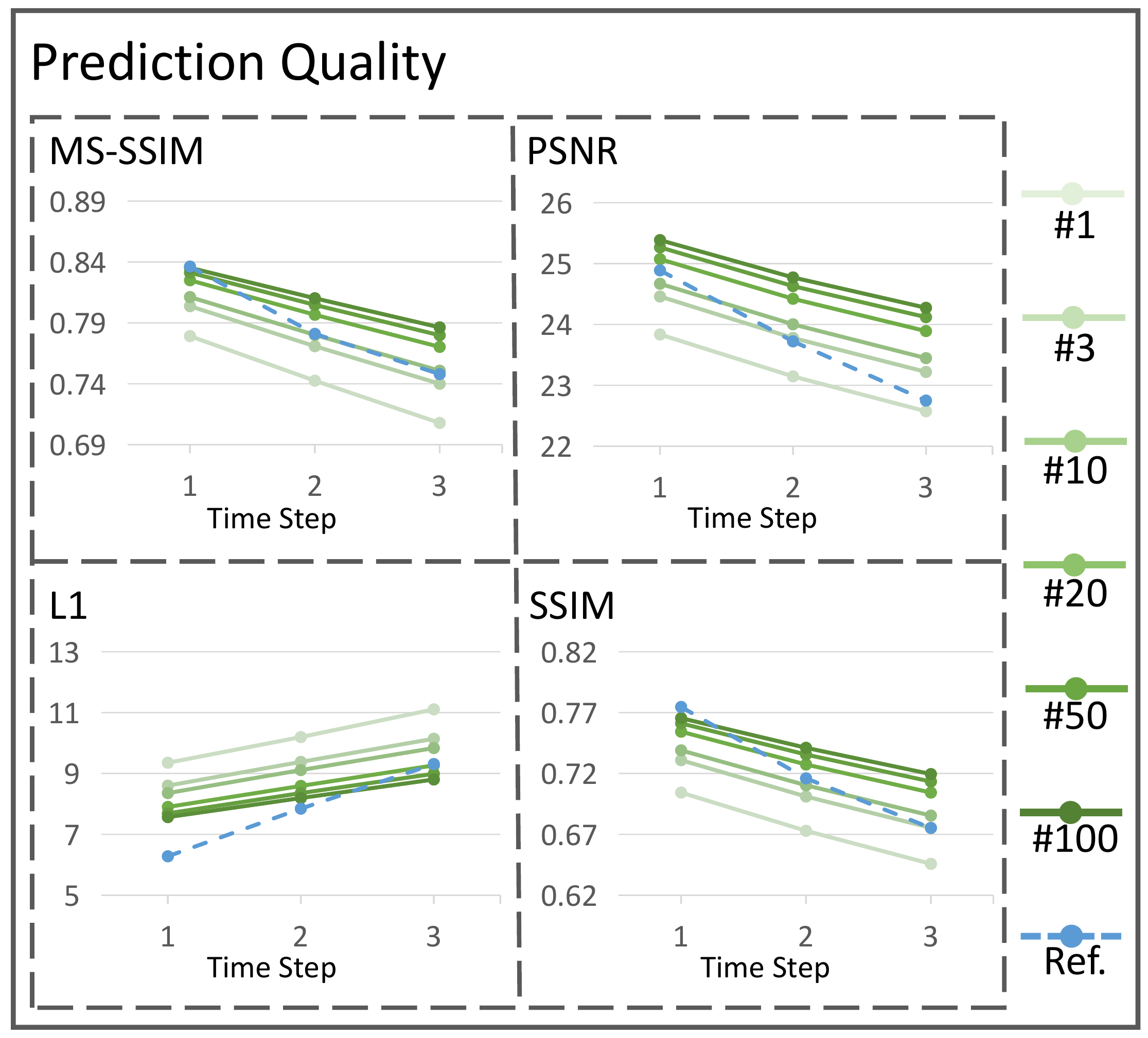}
    \caption{Prediction qualities under different numbers of samples. VAE$^2$ can generate predictions very close to the ground truth future when it is sampled for sufficient times.}
    \vspace{-2mm}
    \label{cityscapes-quality}
\end{figure}
\textbf{Diversity and Quality}.
In order to measure the diversity of the predictions, we make 100 random predictions with each method and compute the PSNR, L1 distance, SSIM, and multi-scale SSIM (MS-SSIM)\cite{wang2003multiscale} between each predicted frame and the ground-truth future frame. In Fig. \ref{cityscapes-variance}, we use box-and-whisker charts to plot the mean (the center of each box), the 3-sigma (solid box), and the 5-sigma (whiskers) of each metric, where sigma denotes the standard deviation. In each sub-figure, different colors represent different methods. It is clear from the figure that the future frames predicted by VAE$^2$ have a larger standard deviation than the reference schemes. It corroborates that VAE$^2$ can predict much more diverse futures than existing methods. Besides, we can find that approaches such as $\beta$-VAE and Anneal-VAE that are designed for model collapse fail in this scenario. They only bring marginal diversity improvement. This implies that we need a specific solution like VAE$^2$ when source data lack stochasticity.  

\begin{figure}
\centering
\includegraphics[width=0.9\linewidth]{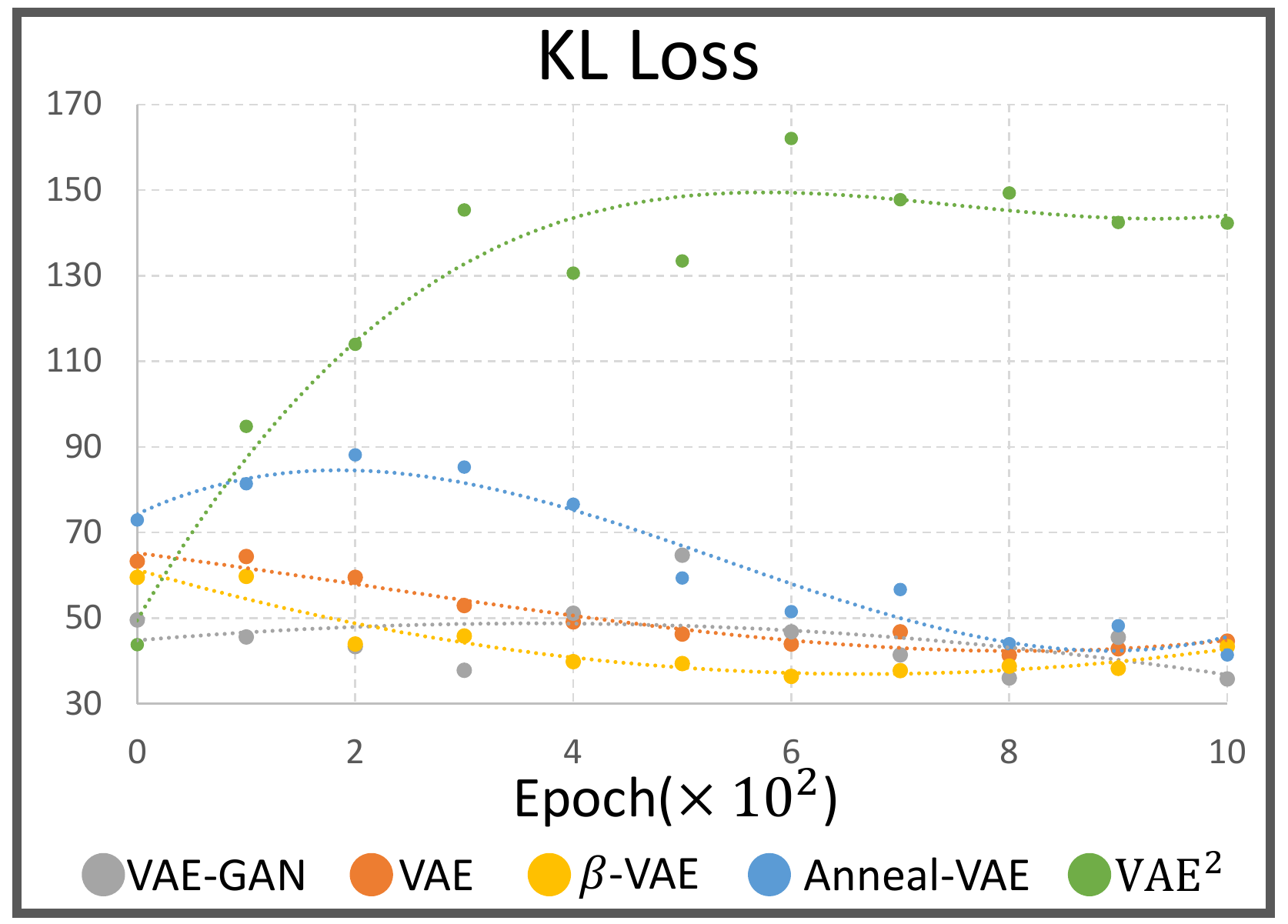}
\caption{KL loss (without normalization) of the latent variable $z$ during training for different schemes. Small KL loss indicates the learned distribution of the hidden variable $z$ is very close to the prior distribution. When KL loss approaches to zero, the latent variable does not encode any useful information.}
\label{z}
\end{figure}

Another intuitive way to check whether the proposed VAE$^2$ helps alleviate the model collapse problem is to evaluate the KL loss of the hidden code $z$. A variational model that collapses to its deterministic counterpart usually has a very small KL loss. As can be viewed in Fig. \ref{z}, the converged KL loss of counterpart methods is significantly smaller than that of VAE$^2$.

Finally, we follow the methods in \cite{babaeizadeh2017stochastic} to evaluate the video prediction quality. In Fig. \ref{cityscapes-quality}, we plot the best prediction among different numbers of samples under various metrics. The reference is a deterministic model\cite{finn2016unsupervised} that directly regresses the ground-truth future. We can see from the figure that as the number of samples increases, the proposed VAE$^2$ can predict video sequences at a quality that is comparable with the reference model in terms of the reconstruction fidelity. As the time-step increases, VAE$^2$ achieves an even better performance than the reference. We also employ Inception Score(IS)\cite{salimans2016improved}, a frequently used no-reference image quality assessment method, to measure the visual quality of the predicted futures. As illustrated by the number beside each legend in Fig. \ref{cityscapes-variance}, the scores are close to each other, which suggests the overall visual quality of the futures predicted by VAE$^2$ are on par with other approaches.

\begin{figure*}
    \centering
    \includegraphics[width=0.95\textwidth]{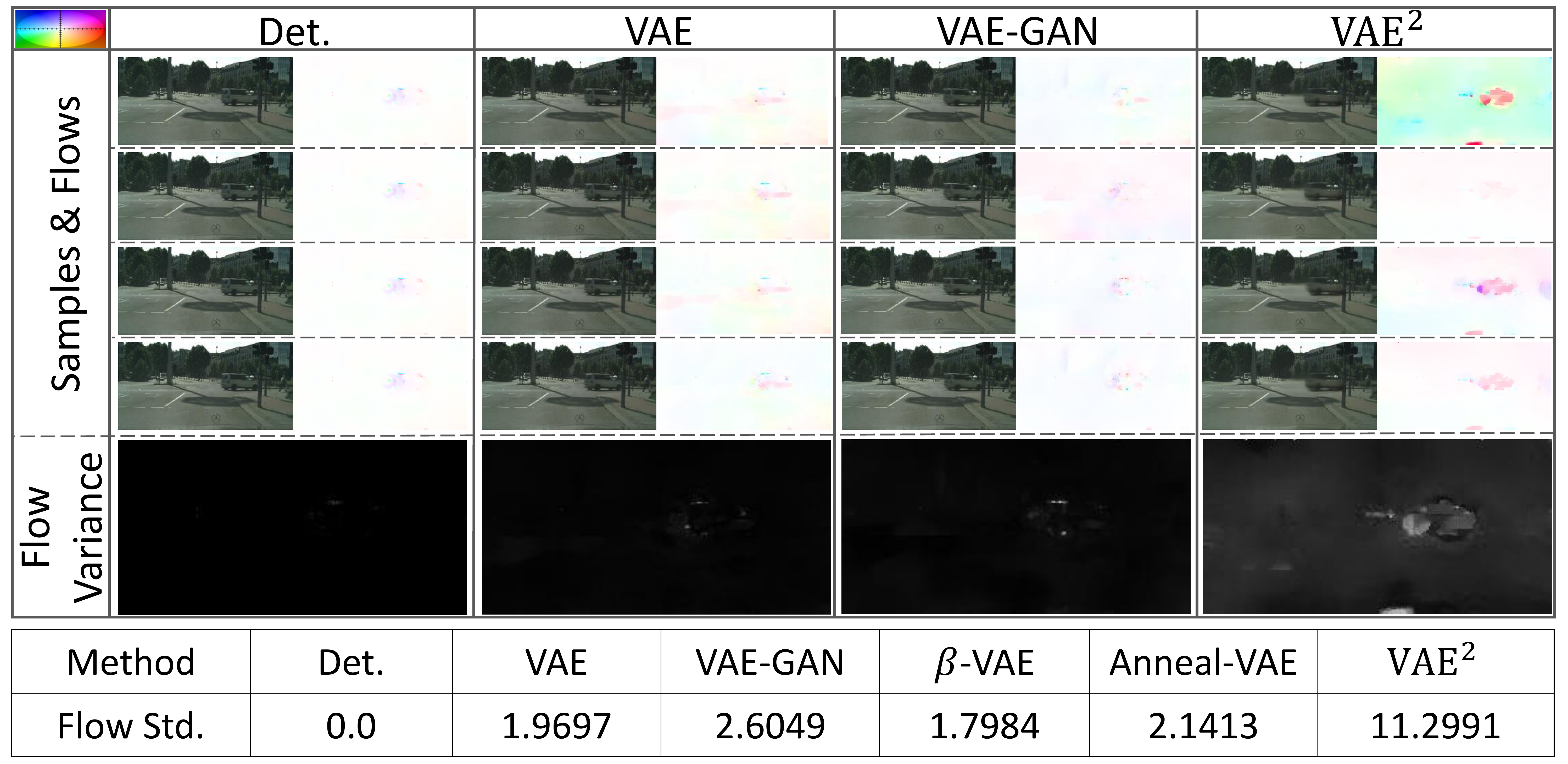}
    \caption{Visualization of predictions and their corresponding optical flow w.r.t. the ground-truth future. We randomly sample four futures with each method. The diverse patterns of the flow images under VAE$^2$ indicate its stochastic behavior. The last row shows the standard deviations of the flow images over 100 samples. The large value of VAE$^2$ supports that it predicts much more diverse future. More visualizations can be found in the appendix.}
    \label{visualization-flow}
\end{figure*}
\textbf{Visualizations}.
Due to limited space, we only visualize single frame predictions with three baseline methods in this section. More visualizations can be found in the appendix. Fig. \ref{visualization} shows three sampled predictions of a single prediction step for each of the schemes to be compared. We notice that the future positions of the fast moving truck predicted by VAE\cite{babaeizadeh2017stochastic} and VAE-GAN\cite{lee2018stochastic} are almost identical to the deterministic baseline. There is little stochasticity among different samples. In contrast, VAE$^2$ achieves noticeable randomness on the moving truck and the moving camera. The predicted camera motion can be observed from the parking car on the right.

\begin{figure}
\centering
\includegraphics[width=1.0\linewidth]{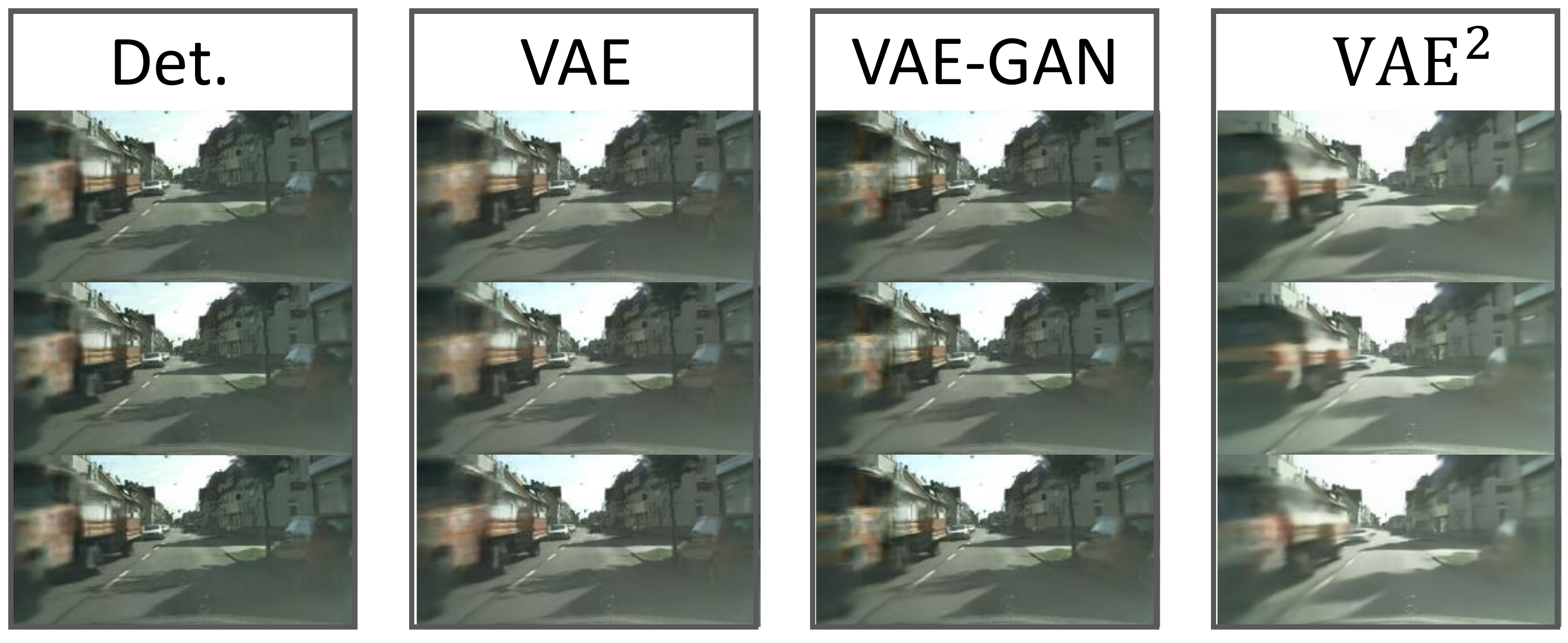}
\caption{Visualization of three predictions for the next frame of an observed video sequence. Each column contains the three predictions made by each method. It can be observed that VAE$^2$ predicts quite different futures while the others make almost identical predictions.}
\label{visualization}
\end{figure}
In order to show the detailed difference among predictions, we compute and colorize\cite{baker2011database} the optical flow between each prediction and the ground-truth future frame. In Fig. \ref{visualization-flow}, we show four predictions of each method and their corresponding optical flow. The color code for optical flows is illustrated at the top-left corner. For example, the first flow map under VAE$^2$ has a green hue, suggesting that the whole background shifts left w.r.t. the ground truth future. This means the camera car is predicted to turn right in the coming future (although it is not the ground-truth future in the dataset). It can also be observed that the displacement patterns of the frames predicted by our method are much more diverse than the other approaches. To quantify such diversity, we compute and visualize the (normalized) standard deviation of the optical flows on 100 different samples for each method. As can be viewed in fifth row, the flow variance of VAE$^2$ has much larger responses on the moving car and the whole background region. The statistics of such diversity on the entire dataset are presented in the table at the last row to illustrate the efficacy of VAE$^2$ on predicting stochastic futures. 


\section{Conclusion and Discussion}
In this paper, we investigate the posterior collapse problem in variational video prediction caused by the videos determinately captured in the wild. We effectively mitigate this problem by explicitly introducing stochasticity into vanilla VAEs and propose an end-to-end framework, VAE$^2$, for optimization. The proposed VAE$^2$ demonstrates its capability of capturing the stochastic information of videos in the wild, which makes variational video prediction more practical in real-world applications. In addtition, we believe that VAE$^2$ can be effectively extended to other sequential prediction problems where training data are lack of stochasticity. We will leave this part to future works. 

We also notice that the inference structure of VAE$^2$ looks similar to the recently proposed two-stage VAE \cite{dai2019diagnosing}. However, this two-stage VAE is designed to address the problem that the hidden code drawn from the encoder is incongruous with the prior, and the first VAE is used to predict the distribution of the hidden code instead of the randomized partial observation as in VAE$^2$.

\bibliography{aaai2021.bib}
\end{document}


\maketitle

This appendix provides the following information, which is not included in the submitted paper due to space limitation.
\begin{itemize}
    \item [-] The detailed proof of eq. (3) and eq. (4) in the paper
    \item [-] Analysis on the tightness of the lower bound.
    \item [-] More details on experimental setup and parameter tuning.
    \item [-] More visualizations.
\end{itemize}
In order to better present the equations involved in our paper, we employ a single-column format for this appendix material.

\section{A Tractable Objective Function}

\subsection{Proof of eq. (3) in the paper}
Assuming that the evolution of a video sequence subjects to a Markov process, where the generation process of a sequence is only conditioned on its previous sequence, we have
\begin{equation}
\begin{split}
    &\log \mathbb{E}_{p(I^s \mid v, I^e)} [p(v \mid I^s) p(I^s \mid I^e)] \\
   =&\log \mathbb{E}_{p(I^s \mid v, I^e)} [p(v \mid I^s, I^e) p(I^s \mid I^e)] \\
   =&\log \mathbb{E}_{p(I^s \mid v, I^e)} [p(v, I^s \mid I^e)] \\
   =&\log \sum_{I^s} p(v, I^s \mid I^e) p(I^s \mid v, I^e) \\
   =&\log \frac{\sum_{I^s} p(v, I^s \mid I^e)p(v, I^s \mid I^e)}{p(v \mid I^e)}.
\end{split}
\label{lowerBound1}
\end{equation}
The Cauchy–Schwarz inequality states that for vectors $u$ and $v$ in an inner product space, it is true that $|\langle u,v \rangle|^2 \leq \langle u,u \rangle \cdot \langle v,v \rangle$, where $\langle \cdot,\cdot \rangle$ is the inner product. Based on this inequality, we have
\begin{equation}
\begin{split}
   \log \frac{\sum_{I^s} p(v, I^s \mid I^e)p(v, I^s \mid I^e)}{p(v \mid I^e)} &\geq \log \frac{\frac{1}{C}[\sum_{I^s} p(v, I^s \mid I^e)]^2}{p(v \mid I^e)} = \log p(v \mid I^e) - \log C, 
\end{split}
\label{lowerBound1}
\end{equation}
where $C$ is a constant. Note that a trivial solution here is to directly employ a vanilla c-VAE to maximize the lower bound $\log p(v \mid I^e)$. However, it is still based on the determinate data pair $(I^e, v)$. By bringing back $I^s$, we derive a non-trivial solution as follows. Let $q_\phi(z \mid I^e, v)$ denote a variational distribution that approximates the true posterior $p(z \mid I^e, v)$ and assuming that the likelihood $p(v \mid I^s, z)$ comes from parametric families of distribution $p_\theta(v \mid I^s, z)$, we have
\begin{equation}
    \begin{split}
        &~~~~~\log p(v \mid I^e) \\
        &= \mathbb{E}_{q_\phi(z \mid I^e, v)}[\, \log p(v \mid I^e) ]\, \\
        &= \mathbb{E}_{q_\phi(z \mid I^e, v)}[\, \log \frac{p(z) p(v \mid I^e, z)}{p(z \mid I^e, v)} ]\, \\
        &= \mathbb{E}_{q_\phi(z \mid I^e, v)}[\, \log \frac{p(z) p(v \mid I^e, z) q_\phi(z \mid I^e, v)}{p(z \mid I^e, v) q_\phi(z \mid I^e, v)} ]\, \\
        &= \mathbb{E}_{q_\phi(z \mid I^e, v)}[\, \log p(v \mid I^e, z) ]\, + \mathbb{E}_{q_\phi(z \mid I^e, v)}[\, \log \frac{p(z)}{q_\phi(z \mid I^e, v)} ]\, + \mathbb{E}_{q_\phi(z \mid I^e, v)}[\, \log \frac{q_\phi(z \mid I^e, v)}{p(z \mid I^e, v)} ]\, \\
        &= \mathbb{E}_{q_\phi(z \mid I^e, v)}[\, \log \sum_{I^s}p(v, I^s \mid I^e, z) ]\, - D_{KL}(q_\phi(z \mid I^e, v) \mid\mid p(z)) \\
        &~~~~~~+ D_{KL}(q_\phi(z \mid I^e, v) \mid\mid p(z \mid I^e, v)) \\
        &\geq \mathbb{E}_{q_\phi(z \mid I^e, v)}[\, \log \sum_{I^s}p(v, I^s \mid I^e, z) ]\, - D_{KL}(q_\phi(z \mid I^e, v) \mid\mid p(z)) \\
        &= \mathbb{E}_{q_\phi(z \mid I^e, v)}[\, \log \sum_{I^s}p(v, \mid I^s, z)p(I^s \mid I^e, z) ]\, - D_{KL}(q_\phi(z \mid I^e, v) \mid\mid p(z)) \\
        &= \mathbb{E}_{q_\phi(z \mid I^e, v)}[\, \log \mathbb{E}_{p(I^s \mid I^e, z)}p_\theta(v, \mid I^s, z) ]\, - D_{KL}(q_\phi(z \mid I^e, v) \mid\mid p(z)).
    \end{split}
\end{equation}
Here $D_{KL}$ denotes the KL divergence and $p(z)$ is the prior distribution of the hidden code $z$. When assuming $p(z)$ to be a centered isotropic multivariate Gaussian $\mathcal{N}(\mathbf{z};\mathbf{0},\mathbf{I})$, where $\mathbf{I}$ denotes the identity covariance matrix, $D_{KL}(q_\phi(z \mid I^e, v) \mid\mid p(z))$ has an analytic form (Please see \cite{kingma2013auto} for more details).

\subsection{Proof of eq. (4) in the paper}
Given a specific hidden random variable $z$ sampled with $q_\phi(z \mid I^e, v)$ in the previous step, let $q_\psi(I^s \mid I^e, z)$ denote a variational distribution that approximates the true posterior $p(I^s \mid I^e, z)$ and assuming that the likelihood $p(I^e \mid I^s, z)$ comes from parametric families of distribution $p_{\theta'}(I^e \mid I^s, z)$. In addition, we follow the assumption in \cite{kingma2013auto} that the generation of data (video sequence in our case) is a random process involving an unobserved random variable, which can be described with $p(I^e) = \int_z p(I^e \mid z)p(z)$. We rewrite $D_{KL}(q_\psi(I^s \mid I^e, z) \mid\mid p(I^s \mid I^e, z))$ as
\begin{equation}\label{eq4}
    \begin{split}
        &~~~~~D_{KL}(q_\psi(I^s \mid I^e, z) \mid\mid p(I^s \mid I^e, z)) \\
        &=\mathbb{E}_{q_\psi(I^s \mid I^e, z)}[\, \log \frac{q_\psi(I^s \mid I^e, z)}{p(I^s \mid I^e, z)}]\, \\
        &=\mathbb{E}_{q_\psi(I^s \mid I^e, z)}[\, \log \frac{q_\psi(I^s \mid I^e, z)}{p(I^s \mid I^e, z)} + \log \frac{p(I^s)p(I^e \mid I^s, z)}{q_\psi(I^s \mid I^e, z)} - \log \frac{p(I^s)p(I^e \mid I^s, z)}{q_\psi(I^s \mid I^e, z)}]\, \\
        &= \mathbb{E}_{q_\psi(I^s \mid I^e, z)}[\, \log \frac{p(I^s)p(I^e \mid I^s, z)}{p(I^s \mid I^e, z)} ]\, - \mathbb{E}_{q_\psi(I^s \mid I^e, z)}[\,\log \frac{p(I^s)p(I^e \mid I^s, z)}{q_\psi(I^s \mid I^e, z)}]\, \\
        &= \mathbb{E}_{q_\psi(I^s \mid I^e, z)}[\, \log p(I^e \mid z)]\, - [\, \mathbb{E}_{q_\psi(I^s \mid I^e, z)}[\,\log p(I^e \mid I^s, z) ]\,  - D_{KL}(q_\psi(I^s \mid I^e, z) \mid\mid p(I^s)) ]\, \\
        &= p(I^e \mid z) - [\, \mathbb{E}_{q_\psi(I^s \mid I^e, z)}[\,\log p_{\theta'}(I^e \mid I^s, z) ]\,  - D_{KL}(q_\psi(I^s \mid I^e, z) \mid\mid p(I^s)) ]\,,
    \end{split}
\end{equation}
where $p(I^e \mid z)$ is the generation process of the video sequence given a specific $z$, which can be regarded as an unknown constant here. Thus, minimizing $D_{KL}(q_\psi(I^s \mid I^e, z) \mid\mid p(I^s \mid I^e, z))$ is equivalent to maximizing $\mathbb{E}_{q_\psi(I^s \mid I^e, z)}[\,\log p_{\theta'}(I^e \mid I^s, z) ]\,  - D_{KL}(q_\psi(I^s \mid I^e, z) \mid\mid p(I^s)) ]\,$, which leads to . (4) in the paper.

\subsection{Tightness Analysis}
 As can be viewed from the Section 1.1 and 1.2 in this appendix, there are two relaxations before we derive the tractable objective function. The first relaxation happens in Eq. (2) which is a standard Cauchy–Schwarz inequality. So its tightness $\epsilon$ is the same as that of the Cauchy–Schwarz inequality. More specifically, we use the difference between the original objective function $\log \frac{\sum_{I^s} p(v, I^s \mid I^e)p(v, I^s \mid I^e)}{p(v \mid I^e)}$ and its lower bound $\log \frac{\frac{1}{C}[\sum_{I^s} p(v, I^s \mid I^e)]^2}{p(v \mid I^e)}$ to measure $\epsilon$ as follows:
 \begin{equation}
    \begin{split}
        \epsilon &= \log \frac{\sum_{I^s} p(v, I^s \mid I^e)p(v, I^s \mid I^e)}{p(v \mid I^e)} - \log \frac{\frac{1}{C}[\sum_{I^s} p(v, I^s \mid I^e)]^2}{p(v \mid I^e)} \\
        &= \log \frac{C \sum_{I^s} p(v, I^s \mid I^e)p(v, I^s \mid I^e)}{p^2(v \mid I^e)} \\
        &\leq \log \frac{C (\sum_{I^s} p(v, I^s \mid I^e))^2}{p^2(v \mid I^e)} = \log{C}. 
    \end{split}
 \end{equation}
 As can be viewed here the margin between the original objective function and its lower bound is no greater than a constant $\log C$. 
 
 The second relaxation happens in Eq. (3) in the appendix. It is a common variational lower bound and its tightness can be directly obtained from the Eq. (3) which is $D_{KL}(q_\phi(z \mid I^e, v) \mid\mid p(z \mid I^e, v))$.

\section{Experimental Setups}
\subsection{Number Sequence Prediction}
\textbf{Backbone Implementations}.
We employ a multi-layer perceptron (MLP) with two hidden layers to instantiate all the necessary encoders and decoders in VAE$^2$, VAE, VAE-GAN, and the deterministic model. The dimensions of the two hidden layers are both 128. We append a non-linear operation ReLU\cite{maas2013rectifier} after each hidden layer.

\textbf{World Model Parameter $\alpha$}.
Parameter $\alpha_k$ used to construct the dataset $D = \{G(\alpha_k, H(\boldsymbol{\hat\epsilon})), k\in[1,K]\}$ is defined by $\alpha_k = 0.1 + 0.0005*(k-1)$. When K is set to $10,000$, $\alpha_k$ ranges from 0.1 to 5.9995. We randomly select 9,000 $\alpha_k$'s for training and the rest 1,000 are for testing.

\textbf{Training Details}.
we use Adam optimizer to train all methods for 1,000 epochs. The weight decay is set to $1e^{-4}$ and the initial learning rate is $1e^{-3}$. The $\lambda$ is set to $0.02$.

\subsection{Cityscapes}
\textbf{Backbone Implementations}.
As discussed in the paper, we employ 18-layer HRNet\cite{sun2019deep} as the backbone network for every encoder, decoder, and discriminator in all methods for fair comparison. HRNet is initially designed for human pose detection and has been successfully extended to many other fields recently \cite{wang2020deep}. HRNet is well-known for its preservation of spatial resolution during inference, which is very beneficial to dense prediction tasks such as semantic segmentation and video prediction. The detailed architecture of the HRNet we used in our paper can be found in \cite{wang2020deep} (referred as HRNetV2-W18). 

Similar to ResNet\cite{he2016deep}, HRNetV2-W18 has a total of four stages. In this paper, for every encoder and decoder which takes the hidden code $z$ as one of the inputs, we concatenate the hidden code to the video sequence's feature maps at the third stage and feed them together into the fourth stage to generate the output.
The dimension of the feature maps from the third stage of HRNetV2-W18 is $72 \times 128 \times 256$ ($channel \times height \times width$) and the dimension of the hidden code $z$ is designed to be $64 \times 1 \times 1$ for all methods in the paper. Therefore we broadcast (repeat) $z$ to a $64 \times 128 \times 256$ map for the concatenation.

The dimensions of video sequences $I^e$, $I^s$, and $v$ are all set to be $3 \times 3 \times 128 \times 256$ ($time \times channel \times height \times width$). In order to make them compatible with 2D CNNs, we combine the time dimension with the channel dimension by reshaping each of the video sequence into a $9 \times 128 \times 256$ tensor. We adjust the kernel dimensions of the first and the last layer in the original HRNetV2-W18 accordingly to fit the dimension changes of the inputs and outputs.

\textbf{Counterpart Method Implementations}
In order to make an apple-to-apple comparison, we replace the backbone networks in different variational video prediction approaches with the same HRNet and remain the original prediction and training framework. We set $\beta$ to $2$ in $\beta$-VAE and employ a sine function ($0$ - $\frac{\pi}{2}$) as the anneal curve in Anneal-VAE. All the other training strategies are kept identical to VAE$^2$. 

\textbf{Training, Inference, and Implementation Details}.

\textbf{Training}. $I^e$ and $v$ are concatenated along the channel dimension and fed into the encoder $\phi$ to generate two $64 \times 1 \times 1$ vectors that represent the mean $\mathbf{m}$ and variance $\mathbf{var}$ of the hidden code $\mathbf{z}$, respectively. The code $z$ is then sampled according to the Gaussian distribution $\mathcal{N}(\mathbf{z}; \mathbf{m}, \mathbf{var})$. The encoder $\psi$ takes as the inputs $z$ and $I^e$ to generate the transition state $I^s$. Then, $I^s$ and $z$ are fed into decoder $\theta^{'}$ and $\theta$ together to reconstruct $I^e$ and $v$ with L1 loss, respectively. Meanwhile, $I^s$ are passed into the discriminator $D_\omega$ to obtain adversarial gradients, where $D_\omega$ is learned by distinguishing the samples generated from the encoder $\psi$ and the samples from the real dataset. Please note that, for the encoder $\psi$, we employ an extra noise signal drawn from a normal distribution which has the same size as $z$ to generate different $I^s$ for the adversarial learning.

\textbf{Inference}. we sample the hidden code $z$ multiple times according to $\mathcal{N}(\mathbf{z};\mathbf{0},\mathbf{I})$. For each sampled $z$, we concatenate it with observation $I^e$ and feed them into encoder $\psi$ to generate $I^s$. Then $I^s$ together with $z$ is fed into decoder $\theta$ to predict the corresponding $v$.

\textbf{Data Augmentation}. we randomly select nine consecutive frames from a video sequence for $I^e$, $I^s$, and $v$, respectively. We resize the spatial dimension to $128 \times 256$ without random cropping or color jittering. 

\textbf{Hyper-parameter}. In order to be comparable with the previous works, the prior distribution for the latent code $z$ is set to be $\mathcal{N}(\mathbf{z};\mathbf{0},\mathbf{I})$ without further tuning. We use Adam optimizer to train all methods for 1,000 epochs on the official train/validation split provided by the Cityscapes. The weight decay is set to $1e^{-4}$ and the initial learning rate is $1e^{-3}$. The $\lambda$ is set to $0.1$, which is decided by the grid search from 0.01 to 1. We observe that $\lambda$ larger than 0.3 makes the generated transition state $I^s$ decay to the partial observation $I^e$, while $\lambda$ smaller than 0.05 makes the $I^s$ decay to the future $v$. We also re-scale the adversarial loss of $D_\omega$ by multiplying $0.1$ to enable a more stable training process. We leverage 8 Nvidia V100 GPUs (32G memory version) and distributed training strategy with batch size 32 and Synchronized Batch Normalization activated. During parameter tuning, we find that extra supervision on $I^s$ at early training stages makes the whole framework converge faster and in a more stable way. Therefore, in addition to the objective function eq. (6) in the paper, we add an L1 loss between $I^s$ generated by the encoder $\psi$ and the ground-truth provided in the dataset with a small coefficient (0.05).

\section{Visualization}
\subsection{Number Sequence Prediction}
We provide more visualizations as a supplement to the Section 4 in the paper.

\begin{figure*}[!htbp]
    \centering
    \includegraphics[width=0.9\textwidth]{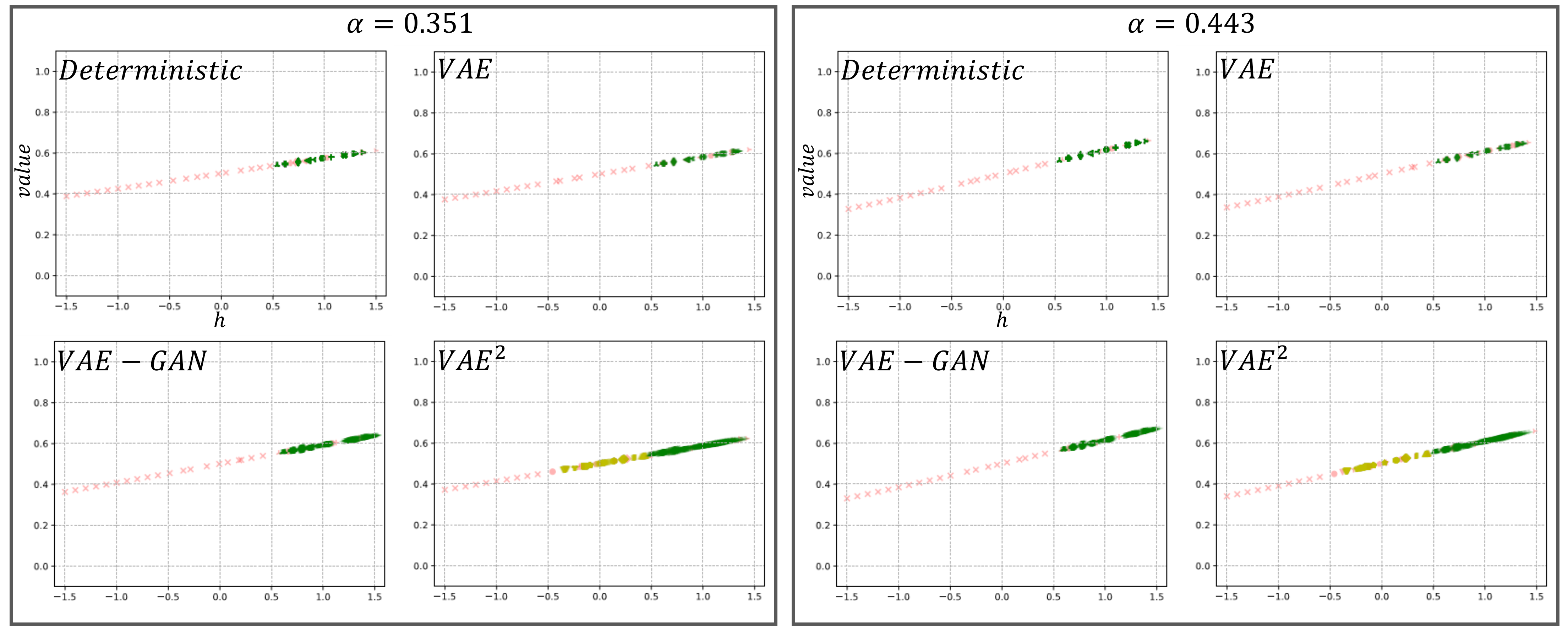}
    \label{appendix_toyexample1}
\end{figure*}
\begin{figure*}[!htbp]
    \centering
    \includegraphics[width=0.9\textwidth]{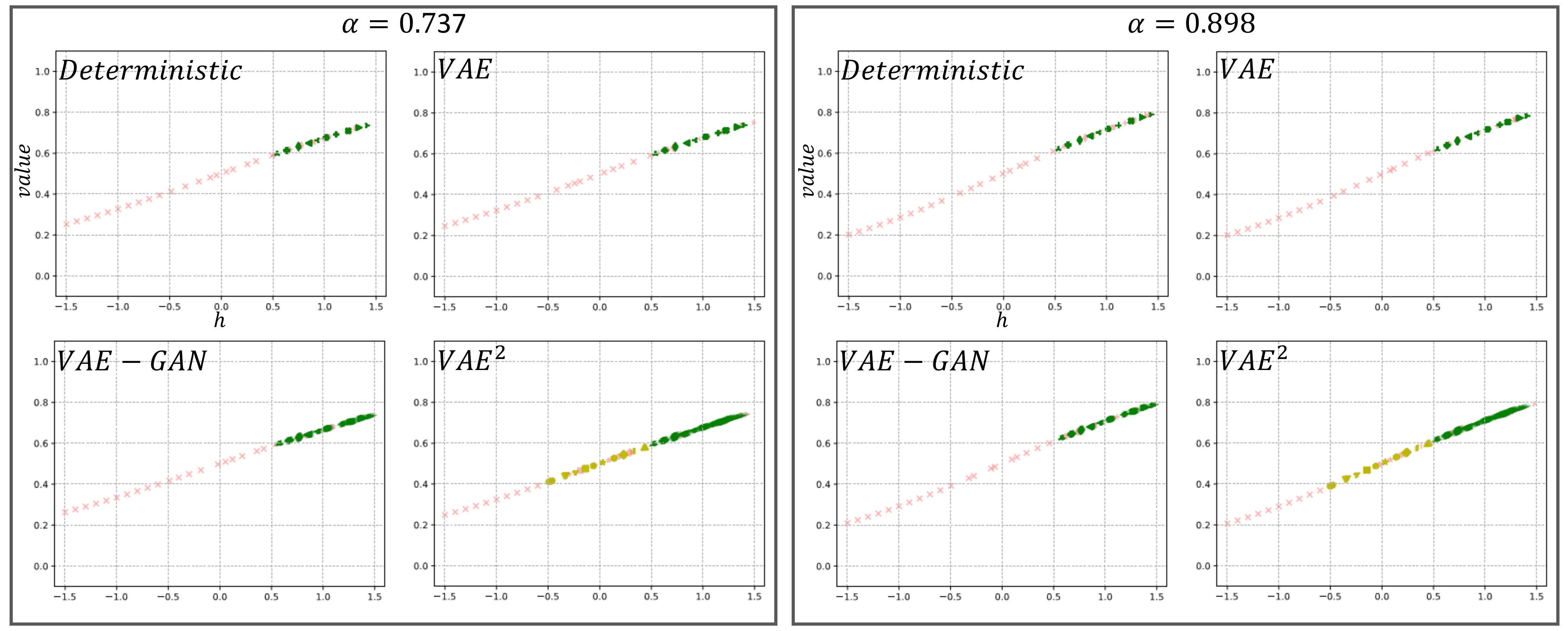}
    \label{appendix_toyexample2}
\end{figure*}
\begin{figure*}[!htbp]
    \centering
    \includegraphics[width=0.9\textwidth]{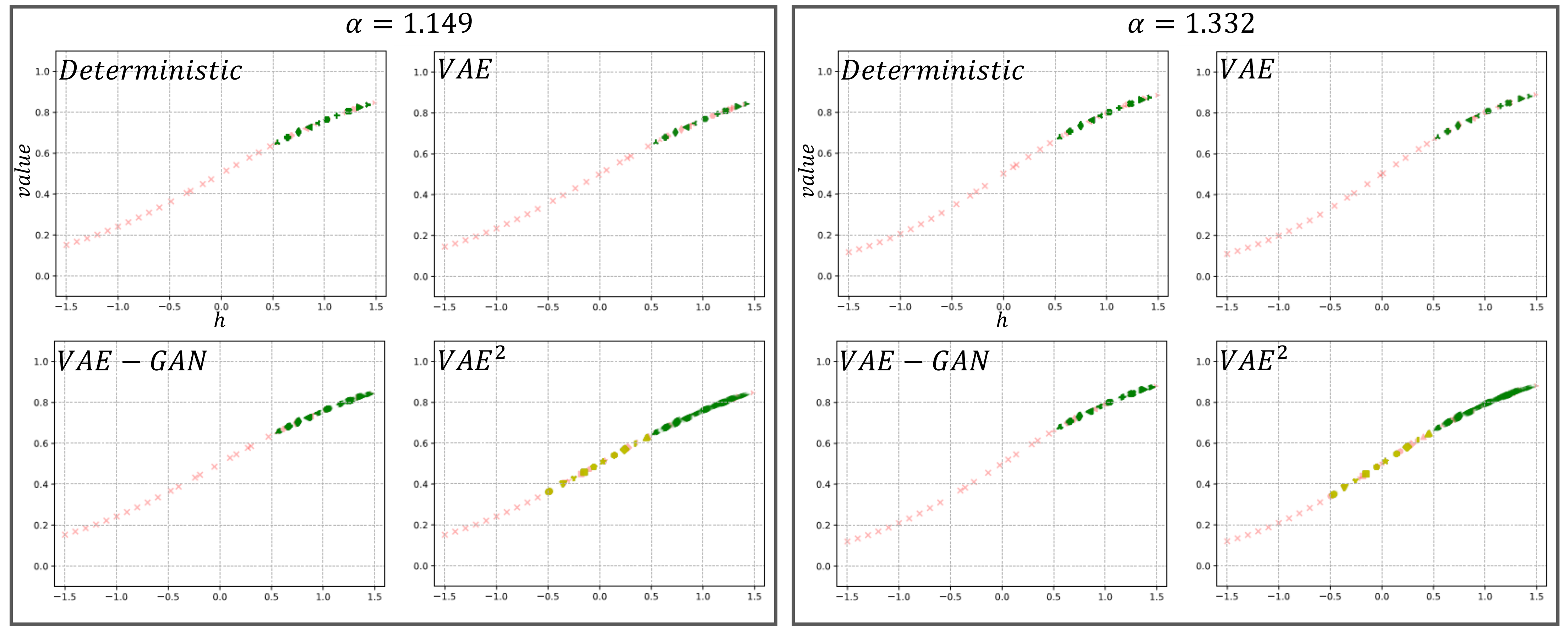}
    \label{appendix_toyexample3}
\end{figure*}
\begin{figure*}[!htbp]
    \centering
    \includegraphics[width=0.9\textwidth]{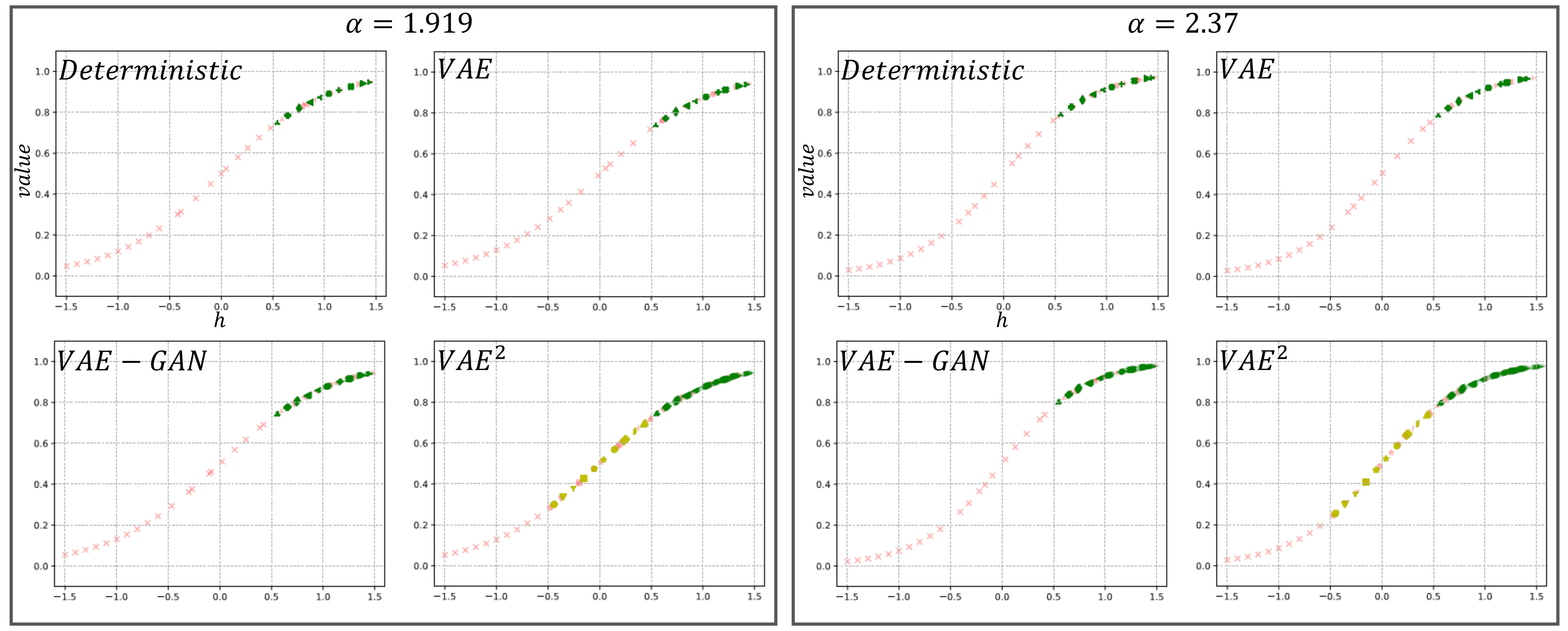}
    \label{appendix_toyexample4}
\end{figure*}
\begin{figure*}[!htbp]
    \centering
    \includegraphics[width=0.9\textwidth]{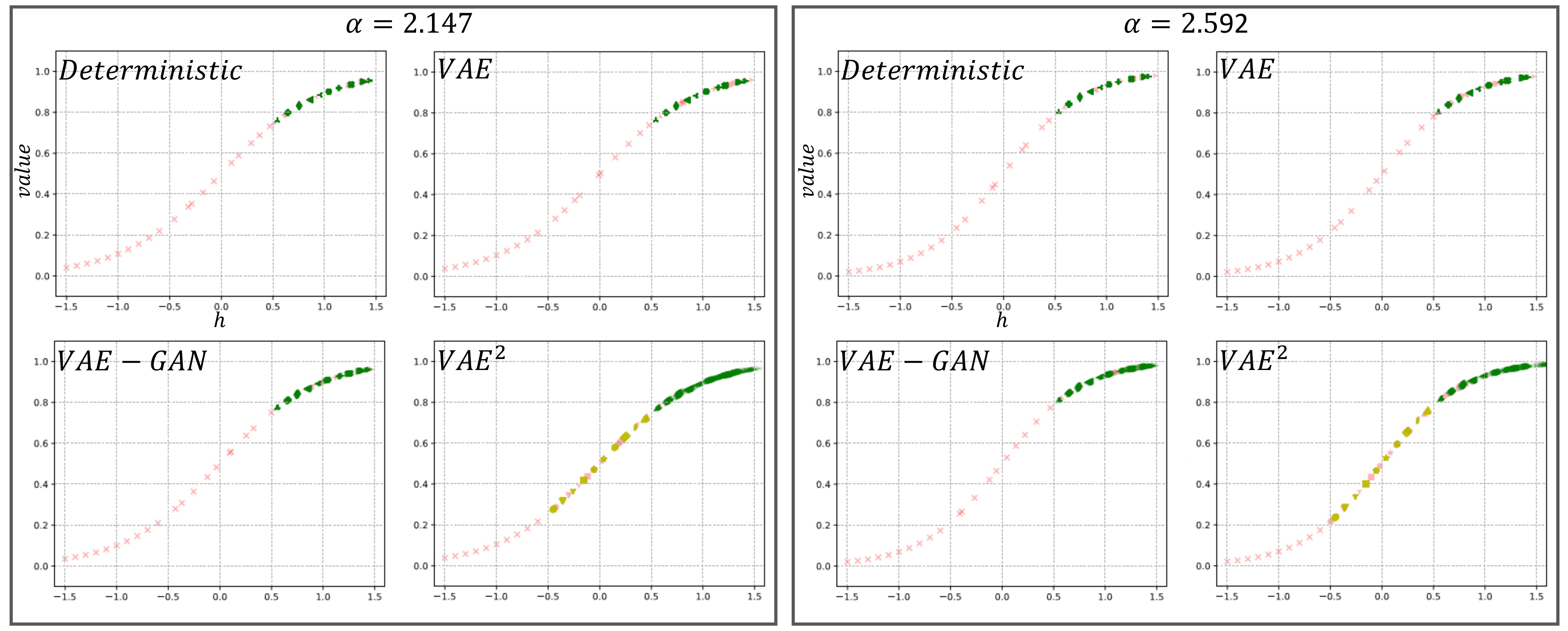}
    \label{appendix_toyexample5}
\end{figure*}
\begin{figure*}[!htbp]
    \centering
    \includegraphics[width=0.9\textwidth]{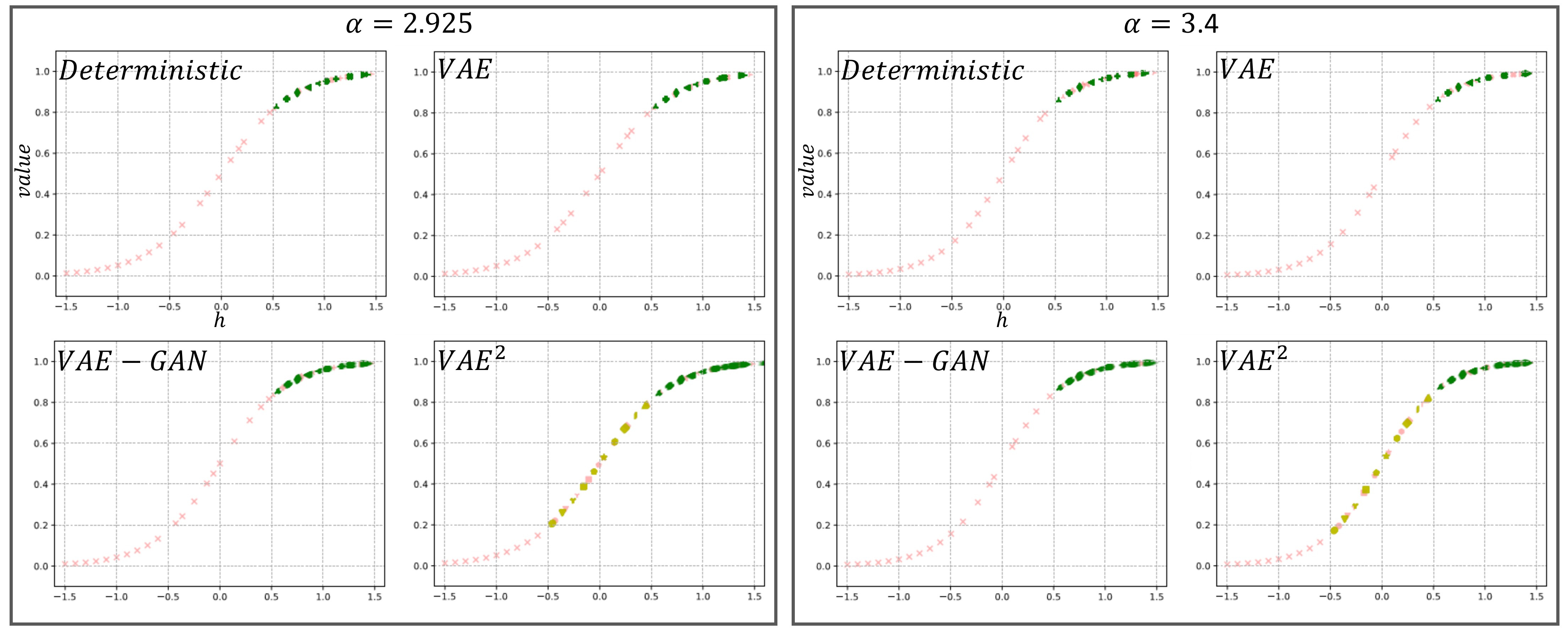}
    \label{appendix_toyexample6}
\end{figure*}
\begin{figure*}[!htbp]
    \centering
    \includegraphics[width=0.9\textwidth]{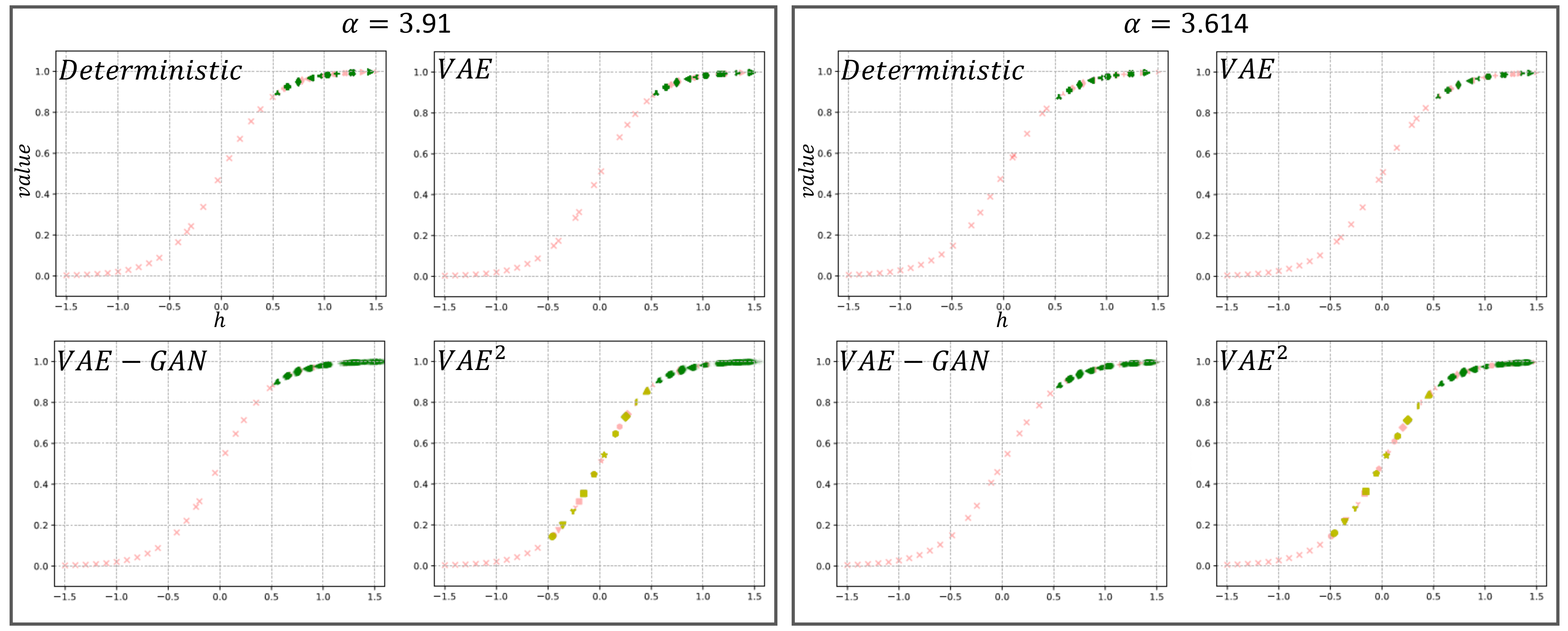}
    \label{appendix_toyexample7}
\end{figure*}
\begin{figure*}[!htbp]
    \centering
    \includegraphics[width=0.9\textwidth]{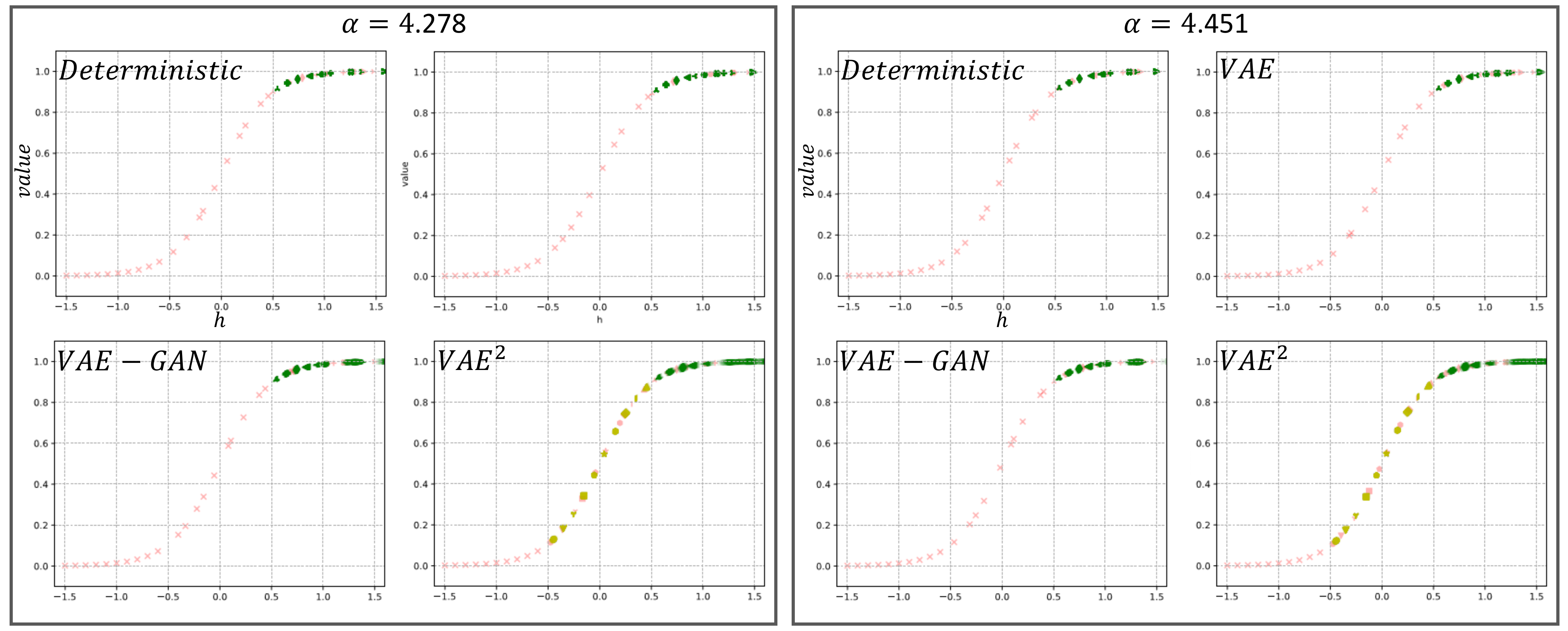}
    \label{appendix_toyexample8}
\end{figure*}
\FloatBarrier

\subsection{Multi-future Predictions on Cityscapes}
We provide more visualizations as a supplement to the Section 5.4 in the paper. The flow variance map is normalized to 0-255, where the highest standard deviation under one visualization group is set to 255. One figure corresponds to one group. The region with higher flow variance indicates the predicted content on this region from different samples is more diverse.

\begin{figure*}[!h]
    \centering
    \includegraphics[width=1.0\textwidth]{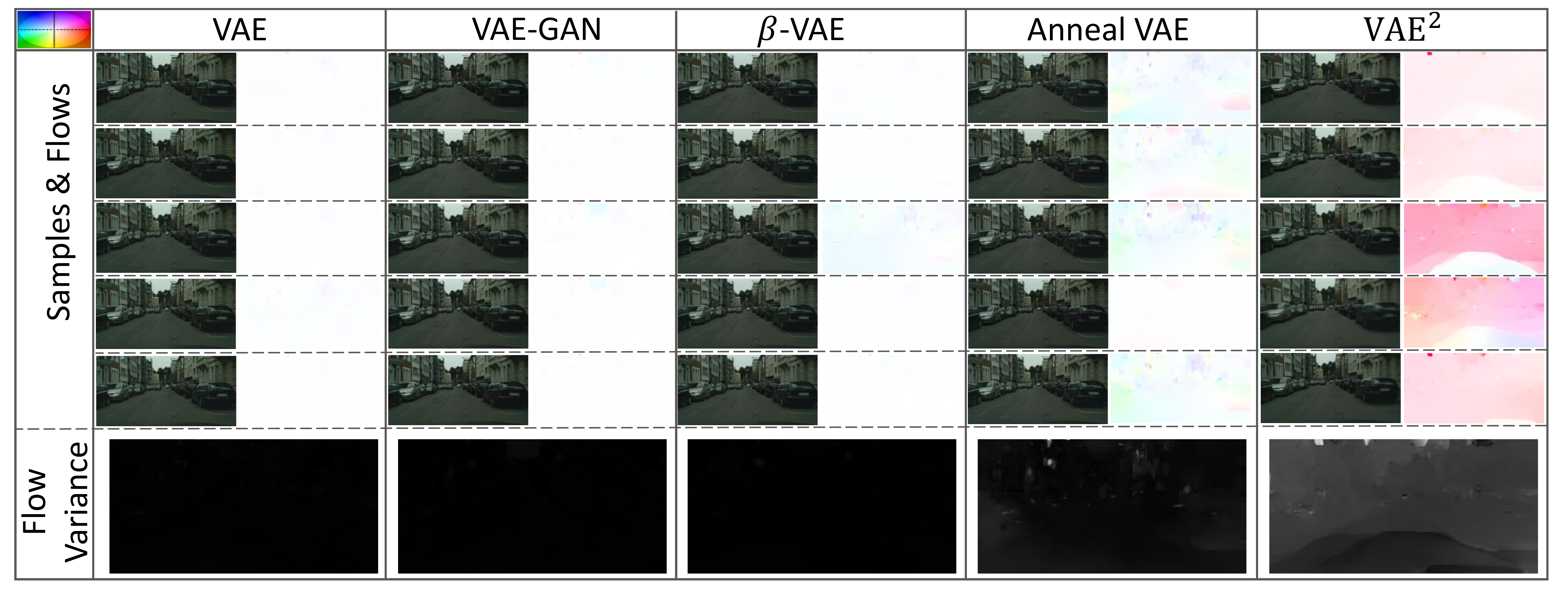}
    \label{appendix_flowvar1}
\end{figure*}
\begin{figure*}[!h]
    \centering
    \includegraphics[width=1.0\textwidth]{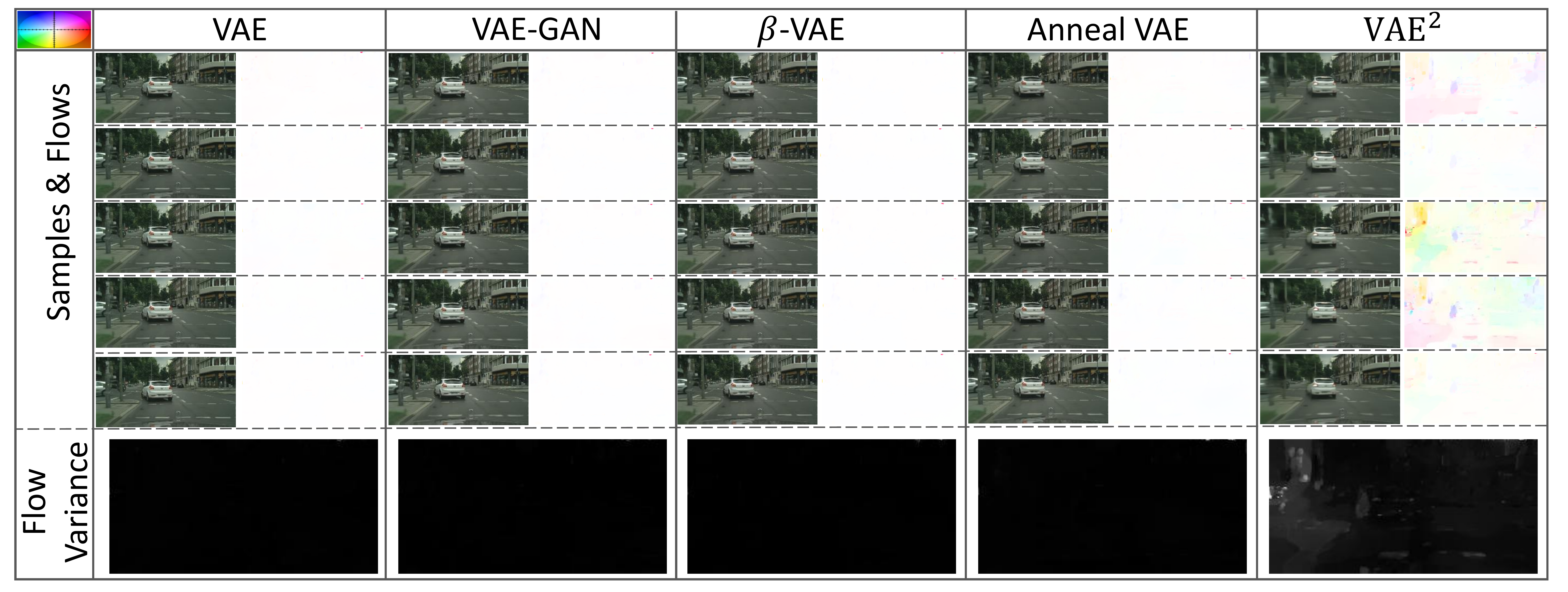}
    \label{appendix_flowvar2}
\end{figure*}
\begin{figure*}[!h]
    \centering
    \includegraphics[width=1.0\textwidth]{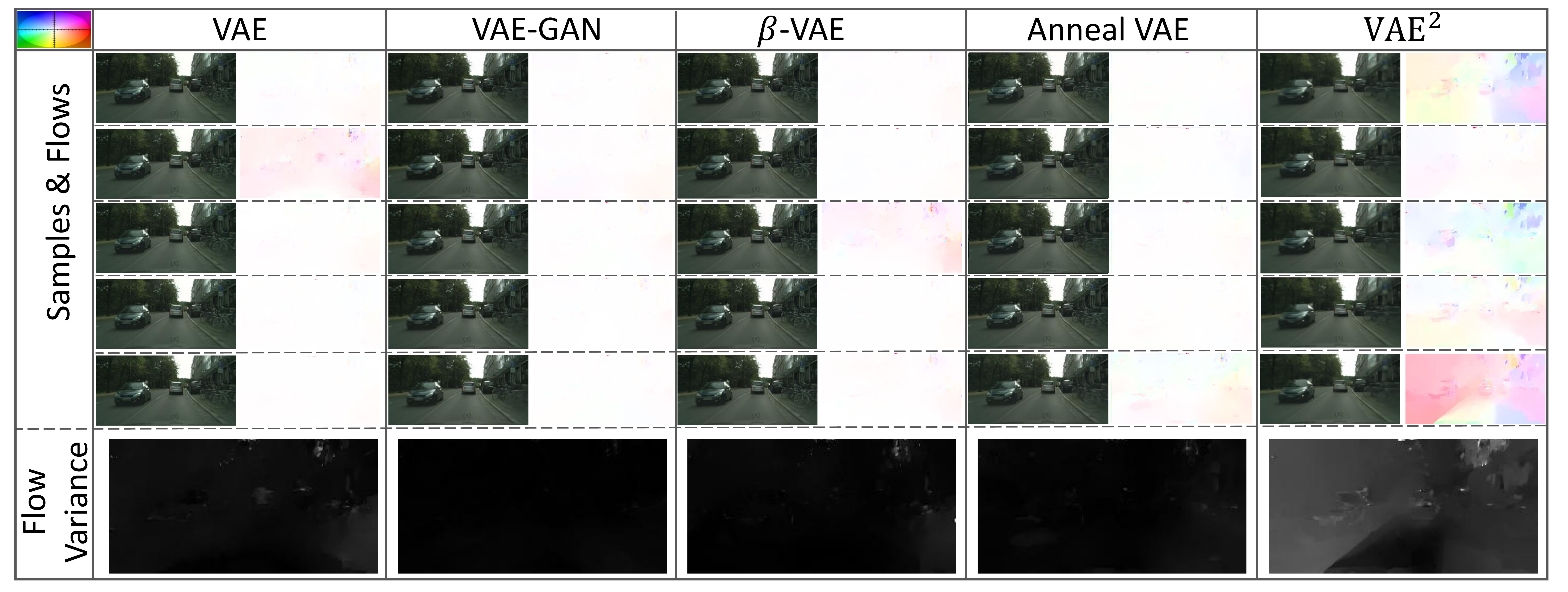}
    \label{appendix_flowvar3}
\end{figure*}
\begin{figure*}[!h]
    \centering
    \includegraphics[width=1.0\textwidth]{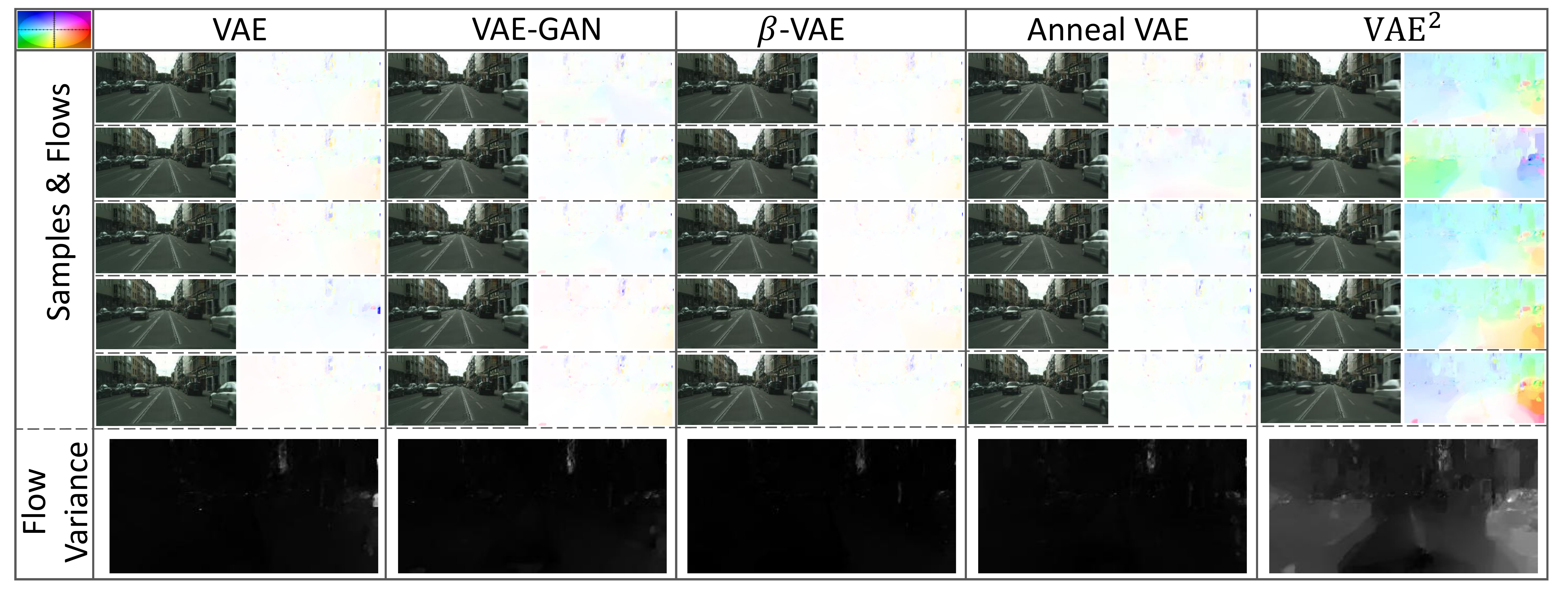}
    \label{appendix_flowvar4}
\end{figure*}
\begin{figure*}[!h]
    \centering
    \includegraphics[width=1.0\textwidth]{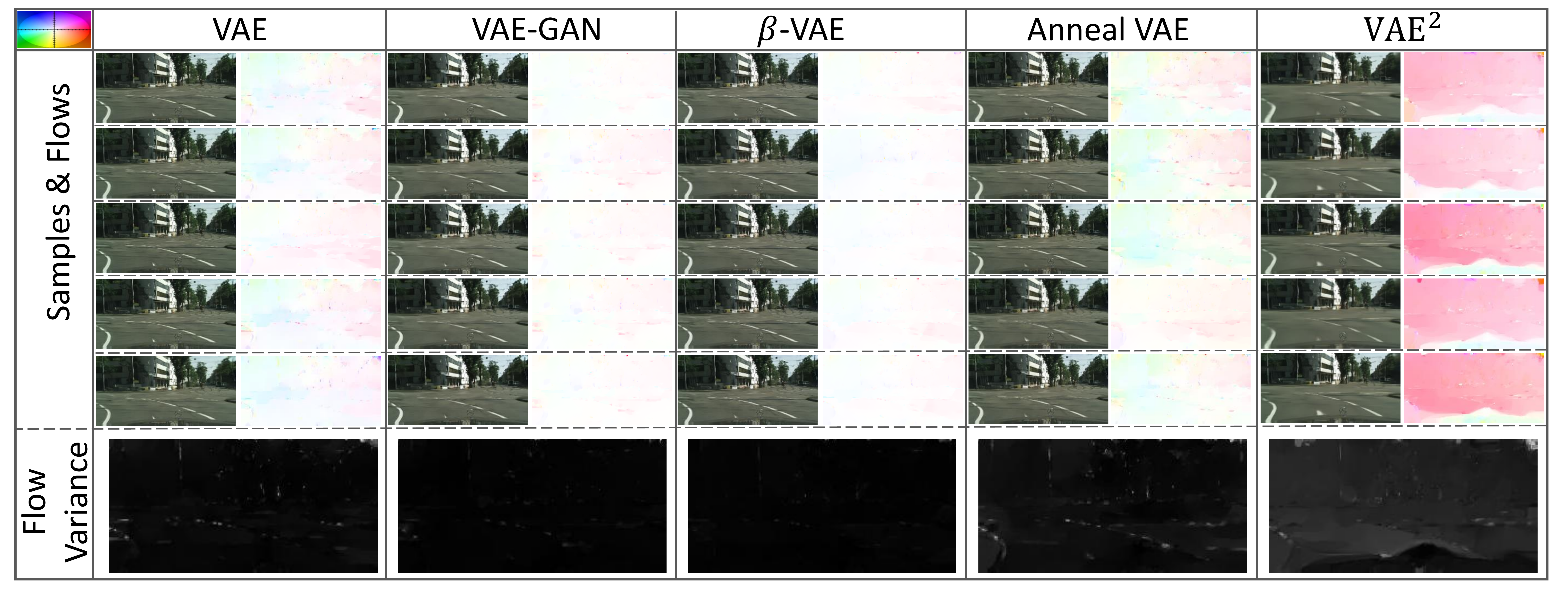}
    \label{appendix_flowvar5}
\end{figure*}
\begin{figure*}[!h]
    \centering
    \includegraphics[width=1.0\textwidth]{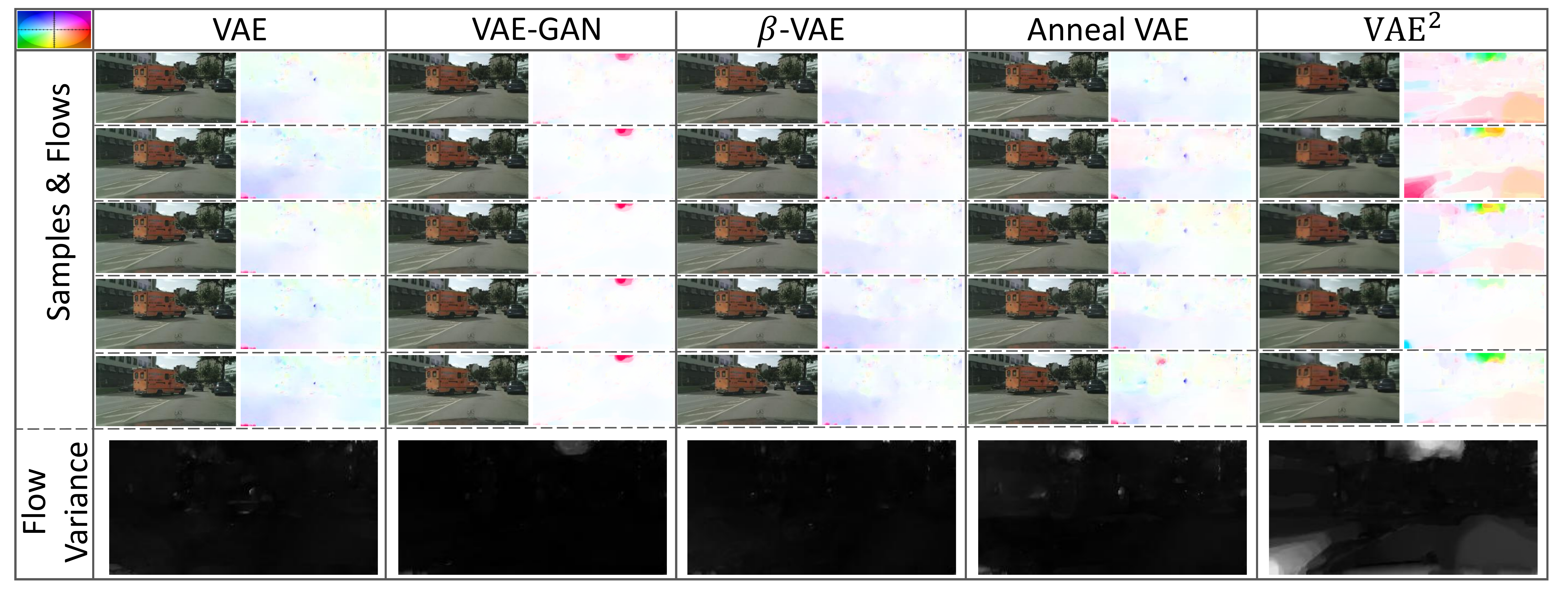}
    \label{appendix_flowvar6}
\end{figure*}
\begin{figure*}[!h]
    \centering
    \includegraphics[width=1.0\textwidth]{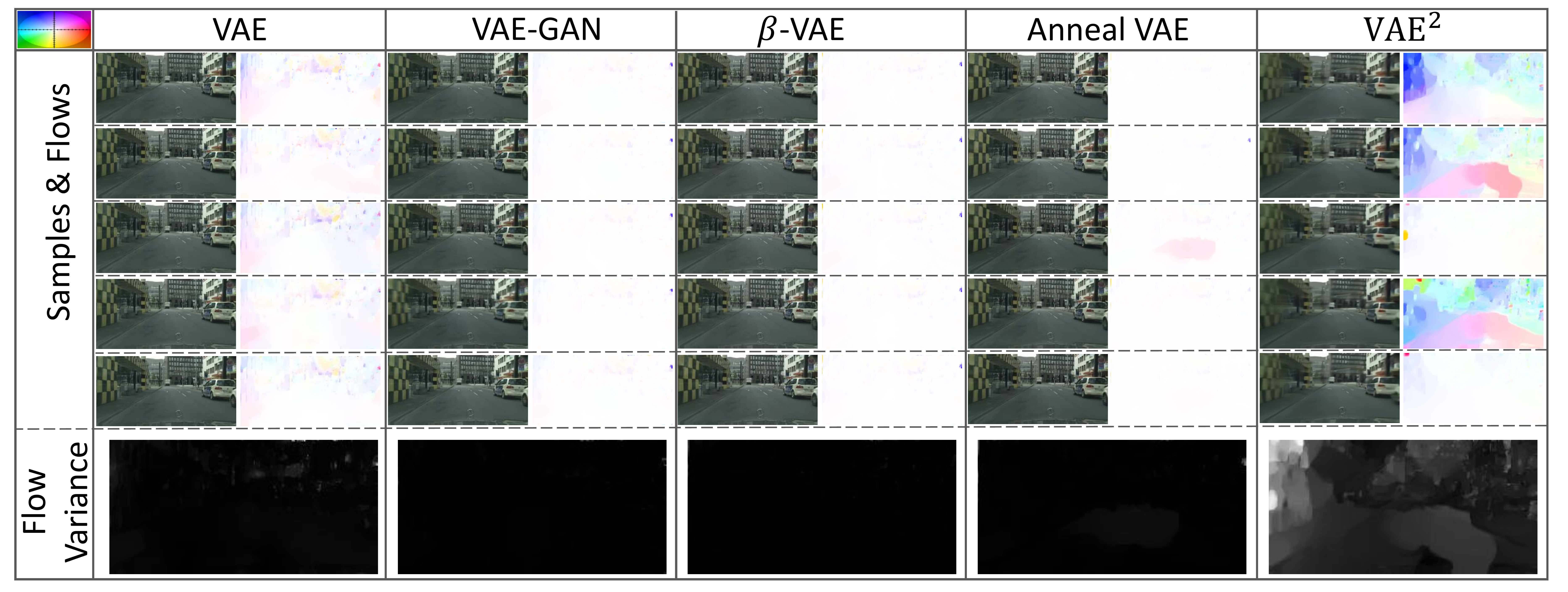}
    \label{appendix_flowvar7}
\end{figure*}
\begin{figure*}[!h]
    \centering
    \includegraphics[width=1.0\textwidth]{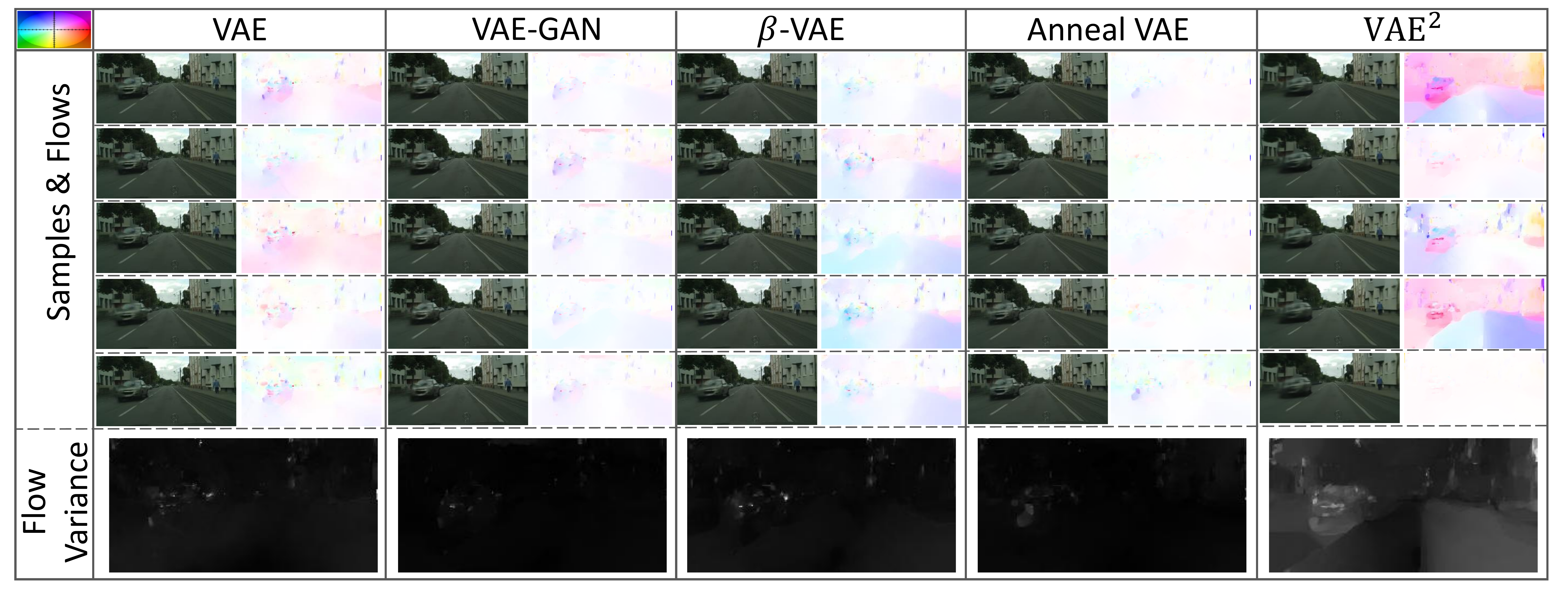}
    \label{appendix_flowvar8}
\end{figure*}
\FloatBarrier

\subsection{Multi-step Predictions on Cityscapes}
We use VAE$^2$ to predict six future frames in an auto-regressive way, where the predicted future in each step is employed as new observation. We randomly select 12 sequences for visualization in the figures below. The orange and green bounding boxes denote the observation and prediction, respectively. As can be viewed from the figures, the futures predicted by VAE$^2$ provide very natural and recognizable content. Here we only visualize one of many possible sequences. In fact, if we choose different predicted futures as new observations, a huge amount different future sequences can be generated. For better illustration, we visualize 100 possibilities of the fourth predicted future in each 
sequence with GIF animation. Please find the corresponding GIF files in the supplementary materials. 

\begin{figure*}[!htb]
    \centering
    \includegraphics[width=1.0\textwidth]{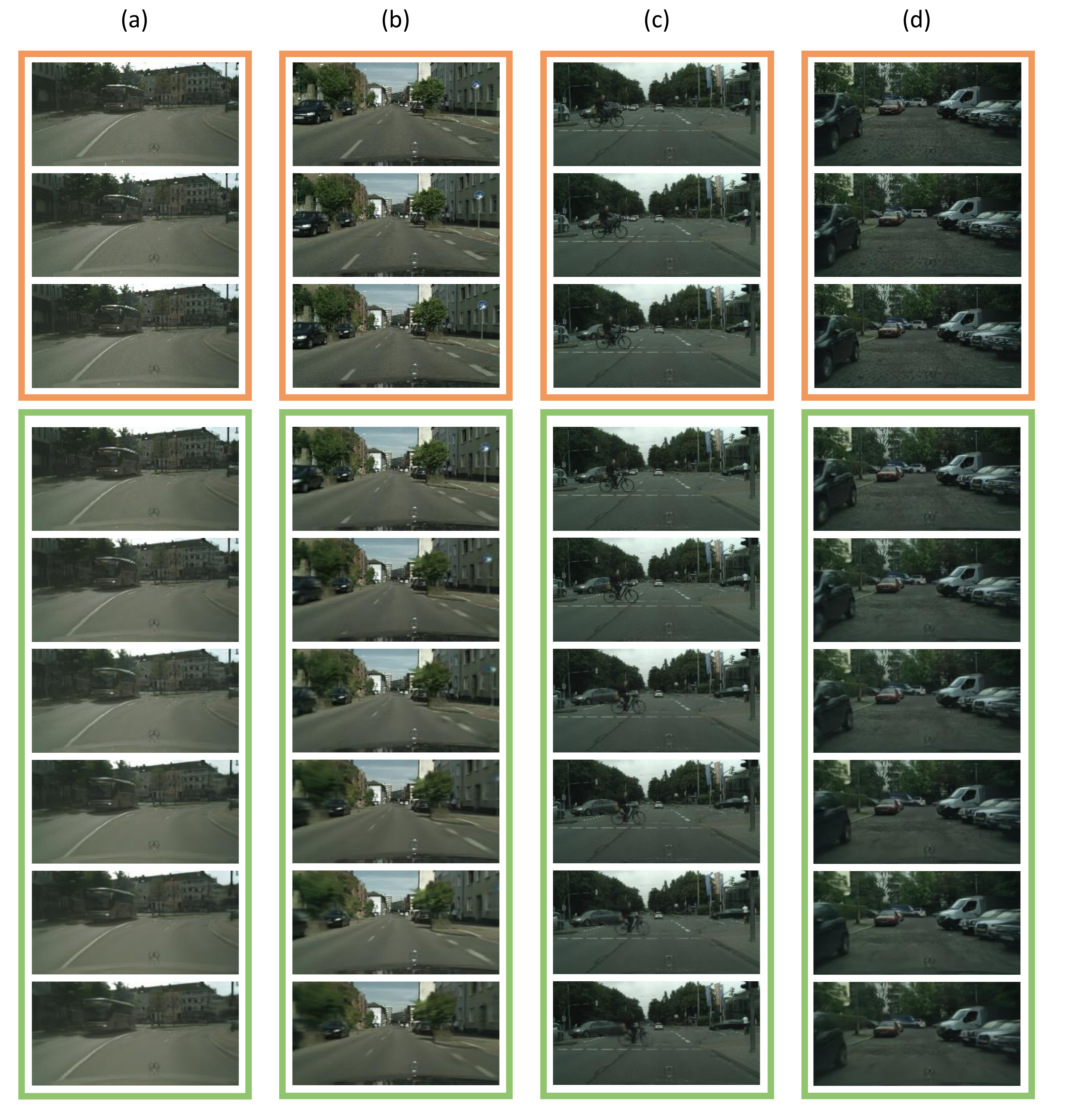}
    \label{appendix_multistep1}
\end{figure*}
\begin{figure*}[!htb]
    \centering
    \includegraphics[width=1.0\textwidth]{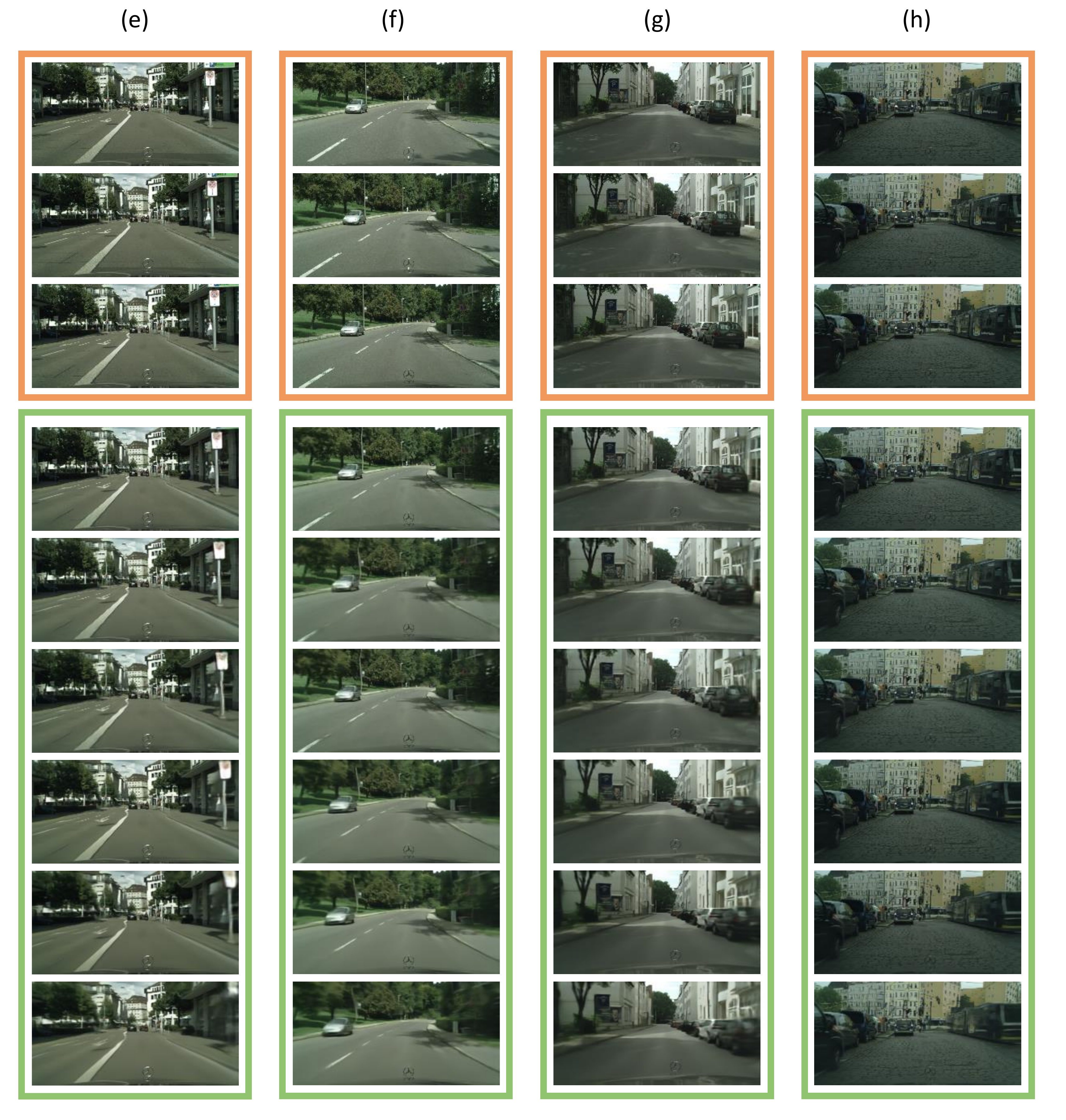}
    \label{appendix_multistep2}
\end{figure*}
\begin{figure*}[!htb]
    \centering
    \includegraphics[width=1.0\textwidth]{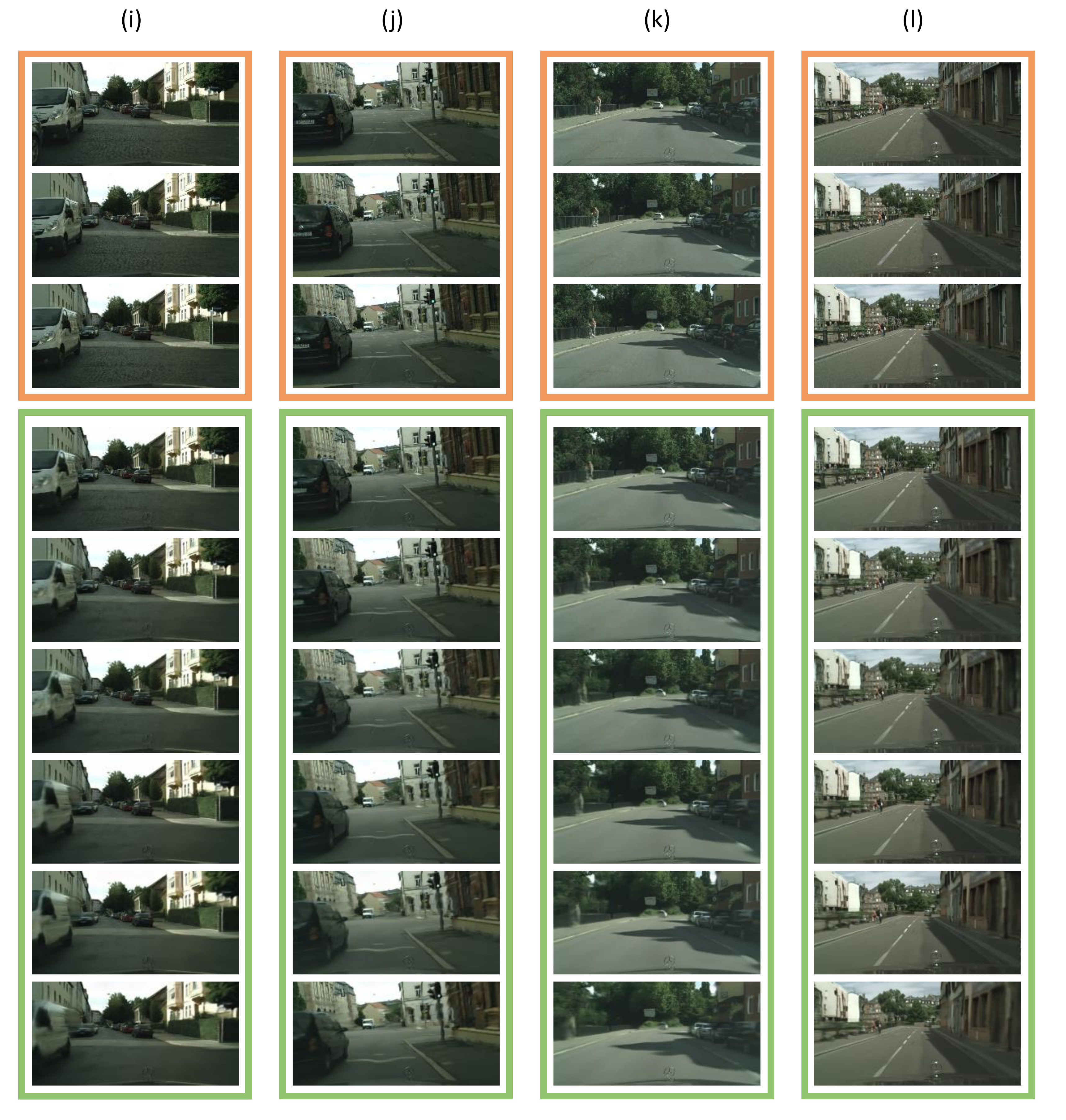}
    \label{appendix_multistep3}
\end{figure*}

\clearpage
{\small
\bibliographystyle{unsrt}
\bibliography{aaai2021}
}